\pdfoutput=1

\documentclass[11pt]{article}

\usepackage[final]{acl}
\usepackage{footmisc}  
\usepackage{caption}
\usepackage{adjustbox}
\usepackage{geometry}
\usepackage{hyperref}  
\usepackage{times}
\usepackage{latexsym}
\usepackage{array}
\usepackage{booktabs} 
\usepackage{multirow} 
\usepackage{refcount}  
\usepackage{todonotes}
\usepackage{menukeys}
\usepackage{tabularx}
\usepackage{slashbox}
\usepackage{amsmath,amssymb}
\usepackage{scrextend}
\usepackage{stfloats}

\usepackage[T2A,T1]{fontenc}
\usepackage[utf8]{inputenc}
\usepackage[russian,english]{babel}
\usepackage{lmodern}
\usepackage{tempora} 

\usepackage{microtype}

\usepackage{inconsolata}

\usepackage{graphicx}

\title{Uncertainty Quantification for Evaluating Gender Bias \\ in Machine Translation}

\author{Ieva Raminta Staliūnaitė\\
  University of Cambridge \\
  \texttt{irs38@cam.ac.uk} \\\And
  Julius Cheng \\
  University of Cambridge \\
  \texttt{jncc3@cam.ac.uk} \\\And
  Andreas Vlachos \\
  University of Cambridge \\
  \texttt{av308@cam.ac.uk} \\}

\begin{document}

\maketitle
\begin{abstract}

The predictive uncertainty of machine translation (MT) models is typically used as a quality estimation proxy. 
In this work, we posit that apart from confidently translating when a single correct translation exists,
models should also maintain uncertainty when the input is ambiguous. 
We use uncertainty to measure gender bias in MT systems. 
When the source sentence includes a lexeme whose gender is not overtly marked, but whose target-language equivalent requires gender specification, the model must infer the appropriate gender from the context and can be susceptible to biases.
Prior work measured bias via gender accuracy, however it cannot be applied to ambiguous cases. 
Using semantic uncertainty, we are able to assess bias when translating both ambiguous and unambiguous source sentences, and find that 
high translation accuracy does not correlate with exhibiting uncertainty appropriately,
and that debiasing affects the two cases differently.\footnote{The code will be available at \url{https://anonymous.4open.science/r/uncertainty_bias_ambiguity-8A4C/}}
\end{abstract}

\section{Introduction}
\label{sec:intro}

Language is inherently ambiguous, and meaning is often resolved through context. However, not all ambiguity is resolvable \citep{van1998ambiguity}.
Preferring one interpretation of an ambiguous statement over another relies on biases \citep{cairns1973effects}.
Natural Language Processing (NLP) models have been shown  not only to reproduce existing social inequalities \citep{beukeboom2013mechanisms} but also to amplify them \citep{dhamala2021bold, tjuatja2024llms}.
Therefore, when faced with an ambiguous input, a well-designed model should remain uncertain to avoid reinforcing biases. 

Uncertainty that arises from unresolvable ambiguity can be understood as a form of aleatoric uncertainty \citep{HORA1996217}, since it is inherent to the data and cannot be reduced. 
Most research considers model uncertainty to be a proxy for prediction correctness \citep{kumar2019verified}, although work has shown that model probability encodes both aleatoric and epistemic uncertainty
\citep{baan2024interpreting}.

Machine Translation (MT) provides an ideal 
testing ground for uncertainty caused by ambiguity. That is because languages differ in the number of grammatical categories they explicitly mark, and translating into a richer language requires disambiguation. This allows for explicitly measuring uncertainty caused by ambiguity by changing the specificity of grammatical gender and comparing the resulting translations. 

\begin{figure}[t]
    \centering
    \includegraphics[width=\linewidth]{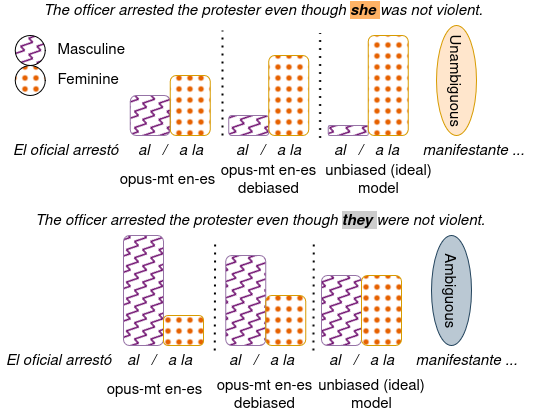}
    \caption{Probabilities for feminine and masculine determiners in a Spanish translation of a sentence containing a noun that is either feminine (referred to as 'she') or ambiguous ('they'), by two existing models and the ideal expected attribution of an unbiased model.}
    \label{fig:illustration}
\end{figure}

In this work, we leverage uncertainty quantification (UQ) to evaluate bias in cases where the model \textit{ought not to be} certain about its predictions due to ambiguity in the input. 
Figure~\ref{fig:illustration} shows two versions of a sentence, with unambiguous (top) or ambiguous (bottom) gender of the noun `\textit{protester}', and the different probabilities assigned to the Spanish translations of the determiner of this noun by an \textsc{Opus-MT} translation model \citep{TiedemannThottingal:EAMT2020}. 
An ideal model should assign a higher probability for the feminine determiner (`\textit{la}') when the gender is disambiguated by the pronoun and produce equal probabilities for masculine and feminine translations when the gender is ambiguous. 
However, in our preliminary analyses, we observed that state-of-the-art MT models, including debiased variants, tend to produce more uniform probability distributions for unambiguous inputs, and less uniform distributions for ambiguous ones.
This indicates that model probabilities are influenced by stereotypical associations between protesting and masculinity, causing the models to default to the masculine form even when no gender preference is warranted (bottom), and predict the feminine form with relatively low confidence despite very clear contextual cues (top).

In our analysis we consider semantic UQ metrics for a number of reasons. First, 
entropy-based uncertainty metrics have been shown to be a suitable measure of ambiguity in human translations \citep{bangalore2016syntactic}, however it has not been applied in MT.
Second, 
quantifying repetition within sampled model outputs, as well as 
using white-box metrics such as 
entropy, has been successfully applied to detecting ambiguous inputs in question answering \citep{cole-etal-2023-selectively, yang2024maqa}.
Third,
semantic UQ metrics
have been shown to capture linguistic nuance in generation tasks, where the diversity of samples is better estimated if their semantics are taken into account 
\citep{cheng2024measuring, farquhar2024detecting}. 

Our goal to detect bias using semantic UQ metrics.
We focus on extrinsic bias \citep{goldfarb-tarrant-etal-2021-intrinsic}, which manifests in model outputs that are affected by stereotypes. 
The currently established metric of gender accuracy evaluates gender bias by measuring how often the model generates the correct gender in the translation \citep{stanovsky2019evaluating}. 
This metric can only be applied in cases where there exists a \textit{correct gender}, i.e. unambiguous instances. 
Bias also appears in ambiguous scenarios, therefore we propose to measure it across the board. 

To study this systematically, we focus on translating sentences from a language that does not mark gender in nouns and verbs (English) into languages that do (Spanish, French, Ukrainian, Russian, German, Italian and Greek). We use the \textsc{WinoMT} \citep{stanovsky2019evaluating} as well as the \textsc{mGeNTe}~\citep{savoldi2025mgente} datasets. 
Both datasets contain instances where the gender is resolvable from context as well as instances where it is not. 
\textsc{mGeNTe} provides alternative translations with gender either explicitly expressed in the target sentence or avoided in the translation. 
\textsc{WinoMT} includes stereotypical gender roles annotations, and we additionally extend it with manual translations. 
We first validate the usefulness of semantic UQ metrics by showing that they correlate well with gender accuracy on unambiguous items across different models. 
Then, we use these metrics to investigate the biases exhibited by the models in ambiguous cases. 
Our main findings are: 
1) the degree of bias correlates with overall model translation accuracy in unambiguous cases,
2) the degree of bias corresponds to translation accuracy at the instance level in ambiguous cases, 
3) debiasing effects vary depending on input ambiguity, translation accuracy, and target language.

\section{Related Work}
\label{sec:related}

Bias in MT has been classified into under-representation and stereotyping, which is persistent even in resolvable settings \citep{savoldi2021gender, ghosh2023chatgpt}.
Various methods for debiasing have been proposed \citep{saunders-byrne-2020-reducing}, however the evaluation of bias is not uniform. 
Earlier work in gender bias in MT has used intrinsic bias evaluation by testing the embeddings of gendered terms and their association with certain topics \citep{ramesh-etal-2021-evaluating}, and extrinsic bias evaluation by counting correct translations against a target \citep{wang-etal-2022-measuring}.
Gender ambiguity in MT has mostly been addressed where it is resolvable \citep{currey-etal-2022-mt, robinson2024mittens}, 
similarly to other types of lexical ambiguity \citep{barua2024using, 10.1162/coli_a_00541}.
Existing solutions for unresolvable ambiguous cases \citep{cho2019measuring, gonen2020automatically, vanmassenhove2021gender, farkas2022measure, piazzolla2023good} rely on tools or human annotation of gender on specific words, limiting their generalisability to other types of ambiguities. 
In contrast, some work circumvents the problem of unresolvable gender ambiguity. 
\citet{lauscher-etal-2023-em} translate the neutral pronoun while preserving the neutrality in the target language, avoiding exhibiting the bias altogether, while \citet{alhafni-etal-2022-user} rewrite gender according to the preferences of the user and \citet{vanmassenhove-etal-2018-getting} include the speaker gender information. 
\citet{stafanovics-etal-2020-mitigating} propose a solution for unresolvable ambiguity by decoupling the task of determining the gender information present in the input from translation, this way circumventing the issue of what should be done when gender information is not present.

When it comes to uncertainty in ML models, methods for distinguishing 
aleatoric and epistemic uncertainty have been proposed \citep{hou2024decomposing}, however they do not distinguish between data randomness and data ambiguity.
Some work has made the link between biases and uncertainty by showing that model uncertainty is linked to the sycophancy bias \citep{sicilia-etal-2025-accounting} as well as the fact that model uncertainty differs when the input mentions different demographic groups \citep{kuzucu2024uncertainty}.
Ambiguity has also been shown to be reflected in uncertainty in different NLP tasks \citep{kim2024aura, cheng2024fairflow},
however, these studies construe uncertainty as a signal for ambiguity stemming from low quality data, not including the ambiguity which is an indispensable feature of language. 

UQ in MT has been used as a proxy for Quality Estimation (QE). For example, \citet{fomicheva2020unsupervised} use MT model uncertainty to estimate translation quality without references, while \citet{glushkova2021uncertainty} apply the same technique to the uncertainty of the QE models themselves. 
Other approaches use UQ to identify difficult instances and enhance training \citep{zhou2020uncertainty, wei2020uncertainty, wu2021uncertainty, zhan2023test}.
\citet{wang2024understanding} examine zero-shot translation and distinguish between model uncertainty and data uncertainty, however their focus with regard to data uncertainty is on noisy, low-quality training data rather than inherent ambiguity. 
\citet{DBLP:conf/icml/OttAGR18} show that uncertainty caused by low quality training data is linked to a deterioration of accuracy in beam search with large beams. 
To the best of our knowledge, no prior UQ-based approach in MT has 
explored ambiguity as a particular type of data uncertainty.

\section{Method}
\label{sec:method}

We propose to quantify gender bias in Neural Machine Translation (NMT) models by characterising how gender is assigned to nouns across the predictive distribution. 
Our bias evaluation method rests on two conditions:

\noindent \textbf{Condition 1:} For source sentences with unambiguous gender, an unbiased model should prefer a translation with correct gender inflection compared to an incorrect inflection. 

\noindent \textbf{Condition 2:} Unbiased models should show higher uncertainty for inputs with ambiguous gender, as compared to unambiguous ones, disregarding the biases in the input.

To do so, we base our approach on recently proposed UQ metrics which are founded on the classic Shannon entropy but take into account similarities between random Monte Carlo
samples from the model. 
Sampling-based metrics are preferred over deterministic decoding methods, as they better capture ambiguity in the ground truth and leverage model stochasticity.
We first provide a brief overview of the UQ methods. 

Let $\mathcal{Y}$ be a random variable whose value is drawn from the predictive distribution of an NMT model $p(y|x)$. Then entropy is defined as: 
\begin{equation}
\mathcal{H}(\mathcal{Y}) = \underset{y \sim \mathcal{Y}}{\mathbb{E}}\left[I(y)\right],
\label{eq:entropy}
\end{equation}
where $I$ is the \textit{surprisal} 
of $y$. In the classic Shannon entropy, $I=-\log p(y)$, but the UQ methods we consider vary in their definition of surprisal.

Semantic Entropy (\textsc{se}) \citep{farquhar2024detecting} is a metric designed to quantify uncertainty by identifying semantic equivalences among output elements. It achieves this by grouping elements \( y \) into clusters \( c \) of semantically equivalent outputs based on a textual entailment model. 
To implement this clustering, we use a multilingual mDeberta model \citep{he2021debertav3} finetuned on the Natural Language Inference (NLI) task \citep{laurer_less_2022}. 
\textsc{se} is intended to indirectly capture clusters of translations based on the gender of the subject or object, given that two identical translations that only differ by gender marking would not entail one another. 

Given the cluster assignment, the surprisal of an output \( y \) under Semantic Entropy is defined as the negative logarithm of the expected probability mass assigned to the cluster \( c \) to which \( y \) belongs:
\begin{equation}
I_{\mathrm{SE}}(y) = -\log \mathbb{E}_{y' \sim \mathcal{Y}} \left[ 1[y' \in c] \right],
\label{eq:surprisal}
\end{equation}
where \( \mathcal{Y} \) is the predictive distribution over outputs and \( 1[\cdot] \) is the indicator function.

Complementing \textsc{se}, Similarity-sensitive Shannon Entropy (\textsc{s3e}) \citep{RICOTTA2006237, cheng2024measuring} generalizes surprisal by incorporating graded semantic similarity between outputs rather than binary clusters, capturing subtler translation differences such as gender marking on verbs and adjectives. Specifically, the surprisal of \( y \) is the negative logarithm of the expected similarity between \( y \) and other outputs \( y' \):
\begin{equation}
I_{\mathrm{S3E}}(y) = -\log \mathbb{E}_{y' \sim \mathcal{Y}} \left[ \mathcal{S}(y, y') \right],
\label{eq:surprisal2}
\end{equation}
where the similarity function \( \mathcal{S} : \mathcal{Y} \times \mathcal{Y} \to [0,1] \) satisfies \( \mathcal{S}(y,y') = 1 \) if and only if \( y = y' \).
Following \citet{cheng2024measuring}, we compute \( \mathcal{S} \) as the cosine similarity between sentence embeddings of \( y \) and \( y' \), generated by a multilingual E5 text embedding model \citep{wang2024multilingual}.

We also define Gender Entropy (\textsc{ge}), which is calculated like \textsc{se} but clusters elements based on the gender class of the translated noun. To determine the gender class, we use
Spacy\footnote{\url{https://spacy.io/}\label{fn:spacy}} and pymorphy2 \citep{korobov2015morphological} morphological parsers. 
\textsc{ge} provides the most direct way to measure uncertainty with regard to the gender. 

To calculate \textsc{se}, \textsc{s3e} and \textsc{ge} we approximate the model's predictive distribution \( p(\cdot \mid x) \) by drawing \( N = 128 \) samples per input using stochastic decoding. Specifically, we perform \(\epsilon\)-sampling \citep{hewitt-etal-2022-truncation} with top-\( p \) sampling (\texttt{epsilon\_cutoff = 0.2}). 
Further details about these UQ methods are given in Appendix~\ref{app:alpha}.

We also compute the log probability of each generated sequence by summing the log probabilities of its tokens. Formally, given a sequence \( s = (w_1, \ldots, w_T) \) and model output logits \(\text{scores}_t\) at step \(t\), we calculate

\[
\log p(s) = \sum_{t=1}^T \log p(w_t \mid w_{<t}, x),
\]

where \(\log p(w_t \mid w_{<t}, x)\) is obtained by applying the log-softmax to \(\text{scores}_t\). This provides a direct likelihood measure of the sequence, used as a simpler alternative to sampling-based surprisal and entropy metrics.

Our gender bias metrics are based on surprisal and entropy from these UQ methods. Thus, to meet \textbf{Condition 1}, in an ambiguous setting, an unbiased model should assign lower surprisal to a translation with the correct gender than to an incorrect one. Accordingly, unbiased models should minimize \textit{relative surprisal}, defined as:
\begin{equation}
\Delta I = \frac{I(y_{\text{correct}}) - I(y_{\text{incorrect}})}{\frac{1}{2}\left(I(y_{\text{correct}}) + I(y_{\text{incorrect}})\right)}.
\label{eq:deltai}
\end{equation}

To meet the \textbf{Condition 2}, unbiased models should show higher uncertainty for an input which is ambiguous with regard to gender, as compared to an input which is unambiguous. It should therefore minimise \textit{relative entropy}, defined as:
\begin{equation}
\Delta\mathcal{H} = \frac{\mathcal{H}(\mathcal{Y}_{\text{unambiguous}}) - \mathcal{H}(\mathcal{Y}_{\text{ambiguous}})}{\frac{1}{2}\left(\mathcal{H}(\mathcal{Y}_{\text{unambiguous}}) + \mathcal{H}\left(\mathcal{Y}_{\text{ambiguous}}\right)\right)}.
\label{eq:relativeentropy}
\end{equation}

\section{Experimental setup}

\paragraph{Data}
\label{sec:data}

To test our proposed bias metrics, an MT dataset containing information about gender ambiguity and stereotypes in the source sentences is required. 
We use \textsc{WinoMT} \citep{stanovsky2019evaluating}, which includes annotations on minimal pairs of 1,584 sentences with masculine, feminine or neutral pronouns referring back to stereotypical or anti-stereotypical gender roles. 
An example of three sentences from the dataset can be seen in Table~\ref{tab:winomt}. The gender of the focus noun `\textit{mechanic}' is unambiguous in the first two sentences based on the contextual information (Pronoun M \& F), but remains ambiguous in the third on account of the neutral (M/F) pronoun. 
The Stereotype (column 3) that mechanics are more often men than women either contradicts (row 1) or aligns with (row 2) the disambiguating context (column 2).\footnote{Please see Appendix~\ref{app:biases} for a broader overview of the biases present in the data.}

\begin{table}[ht]
    \centering
    \scriptsize
    \renewcommand{\arraystretch}{0.9}
    \setlength{\tabcolsep}{2pt}
    \begin{tabular}{p{0.65\columnwidth}cc}
        \hline
        Sentence & Pronoun & Stereotype \\
        \hline
        The \textbf{mechanic} called to inform someone that \textit{he} had completed the repair. & M & M \\
        The \textbf{mechanic} called to inform someone that \textit{she} had completed the repair. & F & M \\
        The \textbf{mechanic} called to inform someone that \textit{they} had completed the repair. & M/F & M \\
        \hline
    \end{tabular}
    \caption{\textsc{WinoMT}~\cite{stanovsky2019evaluating} examples.}
    \label{tab:winomt}
\end{table}

In addition, we experiment with the contemporaneously proposes dataset \textsc{mGeNTe}. The dataset consists of 1,500 sentences, with half of them containing unresolvably ambiguous gender instances, and half where the gender can be determined from the context. While this dataset does not explicitly mark stereotypical gender roles, it provides two alternative translations for each sentence, one with explicit gender marking, and one which avoids specifying the gender in the target language. Table~\ref{tab:mgente} shows examples of both. 

\begin{table*}[ht]
    \centering
    \scriptsize
    \renewcommand{\arraystretch}{0.9}
    \setlength{\tabcolsep}{3pt}
    \begin{tabular}{p{0.25\textwidth}c p{0.33\textwidth} p{0.33\textwidth}}
        \hline
        Sentence & Gender & Translation M/F & Translation G \\
        \hline
        My second comment is that the previous speakers complained that the European Union pays, but the US is listened to. 
        & M/F
        & In secondo luogo, le persone che mi hanno preceduto negli interventi hanno lamentato il fatto che l'Unione europea paga, mentre l'America si fa sentire. 
        & In secondo luogo, gli oratori che mi hanno preceduto hanno lamentato il fatto che l'Unione europea paga, mentre l'America si fa sentire. \\
    \midrule
    The former Commissioner, Mr MacSharry, justified this criminal silence by claiming that the beef sector must not be put at risk. & M & MacSharry, in passato già esponente della Commissione, giustificava in tal modo questo silenzio colpevole, sostenendo che non si doveva mettere a repentaglio il settore bovino. & L'ex Commissario MacSharry giustificava in tal modo questo silenzio colpevole, sostenendo che non si doveva mettere a repentaglio il settore bovino. \\
        \hline
    \end{tabular}
    \caption{\textsc{mGeNTe}~\cite{savoldi2025mgente} examples.}
    \label{tab:mgente}
\end{table*}

The \textsc{WinoMT} dataset has annotations for the focus noun in the input sentence which is ambiguous with regard to gender, whereas in
the \textsc{mGeNTe} the target sentences are annotated for gender on the relavant words. 

\paragraph{Target Languages}
\label{sec:languages}

We selected target languages that express gender through morphological markers on nouns and adjectives, and sometimes verbs, namely Spanish, French, Ukrainian, Russian, Italian, German and Greek. 
The target languages vary in their representation within NLP research.
A relevant idiosyncrasy of the Russian language is that for some nouns describing professions, even if a feminine version exists, it may be considered derogatory in use \citep{komova2024gender}. 
For example, `\textit{\foreignlanguage{russian}{врач}}' is the masculine term for `\textit{doctor}', and the alternative feminine `\textit{\foreignlanguage{russian}{врачиха}}' is considered rude, which leads to `\textit{\foreignlanguage{russian}{врач}}' being used even when the doctor is known to be a woman. 
This results in constructions where masculine nouns are paired with feminine verb forms, or masculine markers are used throughout the sentence.
In order to account for this, we also incorporate a list of professions characterized by (lack of) gender marking in \textsc{WinoMT}, based on the categorisation provided by \citet{komova2024gender}.

\paragraph{Human Translations}
\label{sub:humaneval}

\textsc{WinoMT} does not contain target translations, thus the accuracy of MT models cannot be directly evaluated. To overcome this,
we hired professional translators to translate a set of 100 \textsc{WinoMT} sentences into French, Spanish, Ukrainian and Russian. 
Each sentence is translated twice, with the focus noun in feminine and masculine variants respectively.
They also annotate the translations as Correct or Incorrect with regard to the gender translation in the given context.
For instance, when translating the English sentence ``\textit{The farmer bought a book from the \textbf{writer} and paid her}" into French, where `\textit{writer}' is the focus noun, the feminine `\textit{l'auteure}' should be marked as Correct, while `\textit{l'auteur}' would be Incorrect. 
In ambiguous cases, i.e. if the pronoun in the above sentence was `\textit{they}', both gender translations would be Correct. 
Appendix~\ref{app:translationguidelines} provides the translation guidelines and details, and Appendix~\ref{app:cohen} discusses the quality of human annotations. We release the translations and correctness annotations to the public to enable further research. 

\paragraph{Models}
We experiment with two commonly used translation models, namely \textsc{Opus-MT} \citep{TiedemannThottingal:EAMT2020} and \textsc{m2m100} \citep{fan2021beyond}, as well as a multilingual LLM model \textsc{Tower+} \citep{rei2025tower+}.
\textsc{Opus-MT} models are NMT models trained on freely available parallel corpora. 
\textsc{m2m100} is a many-to-many multilingual translation model, which directly translates between any pair of 100 languages. 
\textsc{Tower+} is an open-weight model which not only exhibits high performance in MT, but also holds strong general capabilities.
To examine how effective debiasing is regarding the expectations stated in Section~\ref{sec:method}, 
we apply the hard-debiasing method from \citet{iluz-etal-2024-applying} on the \textsc{Opus-MT} models, which have been shown to lower bias scores on the \textsc{WinoMT} dataset \citep{stanovsky2019evaluating} while maintaining translation quality.
The hard-debiasing method neutralises the biased words in the representation space, so that neutral words are not associated with a specific gender \citep{bolukbasi2016man}.
We adopt the most effective debiasing approach from \citet{iluz-etal-2024-applying}, which applies debiasing to one-token profession words on the encoder side.
Debiasing methods have been shown to have varying effects depending on the evaluation method \citep{mendelson-belinkov-2021-debiasing, gonen-goldberg-2019-lipstick-pig}, which we also expect to see in our study.
See Appendix~\ref{app:models} for the performance of all models.

\section{Research questions}
\label{sec:experiments}

\paragraph{RQ1: Does Semantic Uncertainty Capture Gender Bias?}
To validate the application of semantic UQ metrics for bias evaluation, we compare their scores with the established gender accuracy metric. 
Gender accuracy uses the morphological parsers described in Section~\ref{sec:method} to determine the focus noun gender in translations. As it relies on gold-standard references, it is applicable only to unambiguous items and unsuitable for cases with multiple valid gender realisations. We therefore limit this experiment to unambiguous items.
We rank all models according to their $\Delta I$ scores and compare this ranking to that based on gender accuracy using Kendall's~$\tau$ and Pearson's~$r$.

\paragraph{RQ2: What does Semantic Entropy Reveal about Bias with Ambiguity?}
To evaluate model bias in ambiguous settings, which has not yet been explored, we compare models using their $\Delta\mathcal{H}$ scores.
To isolate the uncertainty caused by ambiguity from that resulting from poorer model performance, we analyse the relationship between the $\Delta\mathcal{H}$ scores and the translation quality measured by the \textsc{comet} metric \citep{rei-etal-2022-comet}. 
Since the \textsc{WinoMT} dataset does not contain gold target translations, we use the 100 professionally annotated items described in Section~\ref{sub:humaneval}, and the WMT test sets which contain the target languages \citep{ws-2012-statistical, ws-2013-statistical, ws-2014-statistical, wmt-2023, wmt-2024-1} for translation quality evaluation.

\section{Results}
\label{sec:results}

This section presents the results of the bias evaluation using semantic uncertainty metrics.

\paragraph{RQ1: Model Rankings According to Semantic Surprisal and Gender Accuracy Correlate.}
\label{sub:correctness}

\begin{table}[ht]
\centering
\begin{adjustbox}{max width=\columnwidth}
\begin{tabular}{ll|c|c|c|c}
\toprule
\textbf{Lang.} & \textbf{Model} & \textbf{Gender Acc} & \textbf{$\Delta \text{Log prob}$} & \textbf{$\Delta I$ (\textsc{s3e})} & \textbf{\textsc{comet}} \\
\midrule
\multirow{3}{*}{ES} 
& \textsc{Opus-MT}         & 67.95 & 0.00 & -0.10 & 84.90 \\
& deb-\textsc{Opus-MT}     & 68.13 & 0.00 & -0.13 & 84.86 \\
& \textsc{m2m100}          & 70.77 & 0.00 & -0.13 & 84.53 \\
& \textsc{Tower+9B}        & 86.91 & 0.00 & -0.30 & 89.10 \\
\midrule
\multirow{3}{*}{FR} 
& \textsc{Opus-MT}         & 64.27 & 0.01 & -0.04 & 83.56 \\
& deb-\textsc{Opus-MT}     & 64.79 & 0.01 & -0.08 & 83.55 \\
& \textsc{m2m100}          & 61.66 & 0.01 & -0.07 & 83.88 \\
& \textsc{Tower+9B}        & 75.77 & 0.02 & -0.27 & 89.76 \\
\midrule
\multirow{3}{*}{UK} 
& \textsc{Opus-MT}         & 45.34 & 0.00 & -0.03 & 70.79 \\
& deb-\textsc{Opus-MT}     & 46.12 & 0.00 & -0.03 & 70.79 \\
& \textsc{m2m100}          & 47.76 & 0.00 & -0.02 & 76.52 \\
& \textsc{Tower+9B}        & 57.58 & 0.01 & -0.12 & 92.26 \\
\midrule
\multirow{3}{*}{RU} 
& \textsc{Opus-MT}         & 48.57 & 0.00 & 0.00 & 79.37 \\
& deb-\textsc{Opus-MT}     & 48.42 & 0.00 & -0.03 & 79.36 \\
& \textsc{m2m100}          & 48.49 & 0.00 & -0.03 & 80.49 \\
& \textsc{Tower+9B}        & 51.44 & 0.02 & -0.09 & 90.92 \\
\bottomrule
\end{tabular}
\end{adjustbox}
\caption{Gender Accuracy, $\Delta$ Log Probability, and $\Delta I$ (\textsc{s3e}) on Unambiguous instances of \textsc{WinoMT}, \textsc{comet} scores on WMT test sets (see Appendix~\ref{app:models} for details).}
\label{tab:logsurprisalacc}
\end{table}

The results of the first experiment on \textsc{WinoMT} are presented in Table~\ref{tab:logsurprisalacc}. 
We find that while $\Delta \text{Log prob}$ does not correlate with the gender accuracy ranking, $\Delta I$ (\textsc{s3e}) shows a statistically significant negative correlation with gender accuracy (Spearman's \( \rho = -0.84 \); see Appendix~\ref{app:kendalpearson}). 
We attribute the effectiveness of this metric in distinguishing Correct from Incorrect gender translations to its ability to capture gender information beyond noun morphology, including verb inflections and agreement, through its embedding representations.
The flexibility of \textsc{s3e} enables it to encode nuances that \textsc{ge} does not.
For example, when translating a sentence with a feminine pronoun into Russian, \textsc{Opus-MT} generates sentences with a masculine noun and a verb that is either feminine or masculine (e.g., ``\foreignlanguage{russian}{\textit{Перевозчик поблагодарила}}~({fem}) / \foreignlanguage{russian}{\textit{поблагодарил}}~({masc})" ``\textit{The courier thanked}").
This is reflected in a higher $\mathcal{H}$ score assigned to this sentence by \textsc{s3e}: $\mathcal{H}$ = 0.65, than \textsc{ge}: $\mathcal{H}$ = 0.00, as the variation in verb inflections is only captured by \textsc{s3e}. 

The very strong negative correlation between $\Delta I$ (\textsc{s3e}) and gender accuracy thus validates the core component of our proposed metric for evaluating bias in machine translation.
Our findings indicate that we can rank models according to their bias in unambiguous settings with our proposed metric as well as it can be done with gender accuracy, but without the need for explicit gender annotations. 

We do not find a correlation between the surprisal scores and gender accuracy in the \textsc{mGeNTe} dataset (please see Appendix~\ref{app:mgenteresults} for the scores).
There are a number of factors that explain this. 
First,
in \textsc{WinoMT} the Incorrect translations are explicitly the wrong gender, while in \textsc{mGeNTe} the Incorrect translations contain neutral expressions which do not match the explicit gender markers in the remainder of the same sentence (please see Translation N in the top example in Table~\ref{tab:mgente}). This is not strictly grammatically incorrect, only pragmatically less preferred, which is expected to elicit less surprisal from the model.
Second, 
the \textsc{mGeNTe} dataset is not designed to include bias via stereotypes, as the gendered content of the sentence neither strongly aligns with nor contradicts the gender marking.
This leads to smaller differences between the predicted likelihood of different gender expressions, with the highest absolute value of  $\Delta H$ reaching 0.25, compared to 0.74 on \textsc{WinoMT}.

\paragraph{RQ2: Translation Accuracy Affects the Bias–Entropy Relationship Differently Across Levels of Analysis.}
\label{sub:entropy}

Having validated the \textsc{s3e} metric in the first experiment, we investigate the results of $\Delta\mathcal{H}$ (\textsc{s3e}) as a bias metric.  
Table~\ref{tab:s3e_entropy_structured_sorted} illustrates that all models exhibit the desired negative $\Delta\mathcal{H}$ for some languages in the \textsc{WinoMT} dataset.
\textsc{Tower+9B}, which performs best in translation accuracy, also exhibits the desired unbiased behavior with consistently negative $\Delta\mathcal{H}$, suggesting that better models are less biased. 
Surprisingly, when it comes to ambiguous instances, contrary to the unambiguous cases, the languages where the models perform better on translation accuracy, namely Spanish and French, are not generally less gender biased under ambiguity (model rankings according to all metrics are available in Appendix~\ref{app:rankings}).
Nonetheless, we observe the expected effect of debiasing on the Spanish and French translations (lower $\Delta\mathcal{H}$ for deb-\textsc{Opus-MT}). 

\begin{table}[ht]
\centering
\footnotesize
\begin{tabular}{@{}l l | ccc@{}}
\toprule
Lang. & Model & Unamb & Amb & $\Delta\mathcal{H}$\\
\midrule
 \multirow{3}{*}{ES} 
    & \textsc{Opus-MT} & 1.23 & 1.12 & 0.09\\
    & deb-\textsc{Opus-MT} & 0.97 & 0.89 & 0.08\\
    & \textsc{m2m100} & 1.79 & 1.45 & 0.19 \\
    & \textsc{Tower+9B} & 1.67 & 1.82 &  \textbf{-0.15} \\
\midrule
\multirow{3}{*}{FR} 
    & \textsc{Opus-MT} & 1.79 & 1.43 & 0.20\\
    & deb-\textsc{Opus-MT} & 1.21 & 1.08 & 0.11\\
    & \textsc{m2m100} & 3.22 & 2.78 & 0.14\\
    & \textsc{Tower+9B} & 2.72 & 3.46 & \textbf{-0.74} \\
\midrule
\multirow{3}{*}{UK} 
    & \textsc{Opus-MT} & 1.96 & 2.16 & \textbf{-0.10}\\
    & deb-\textsc{Opus-MT} & 1.98 & 2.15 & \textbf{-0.09}\\
    & \textsc{m2m100} & 2.05 & 2.28 & \textbf{-0.11}\\
    & \textsc{Tower+9B} & 4.12 & 4.85 & \textbf{-0.73} \\
\midrule
\multirow{3}{*}{RU} 
    & \textsc{Opus-MT} & 1.56 & 1.68 & -0.08\\
    & deb-\textsc{Opus-MT} & 1.05 & 0.97 & 0.08\\
    & \textsc{m2m100} & 1.83 & 2.29 & \textbf{-0.25}\\
    & \textsc{Tower+9B} & 3.15 & 3.63 & \textbf{-0.48} \\
\bottomrule
\end{tabular}
\caption{Unambiguous and Ambiguous $\mathcal{H}$ (\textsc{s3e}) on \textsc{WinoMT}. Statistically significant correlations between $\Delta\mathcal{H}$ and \textsc{comet} scores on the instance level are in \textbf{bold}.}
\label{tab:s3e_entropy_structured_sorted}
\end{table}

The results on the \textsc{mGeNTe} dataset corroborate the finding that the best performing model (\textsc{Tower+9B}) exhibits the least amount of bias (see Table~\ref{tab:s3e_entropy_structured_sorted2} in Appendix~\ref{app:mgenteresults}).
The effects are smaller compared to the \textsc{WinoMT} dataset, and debiasing does not show an effect on \textsc{mGeNTe}, which may be explained by the fact that the dataset is not explicitly stereotypical and less bias can be observed. 

\begin{figure*}[!htbp]
    \centering
    \includegraphics[width=0.843\linewidth]{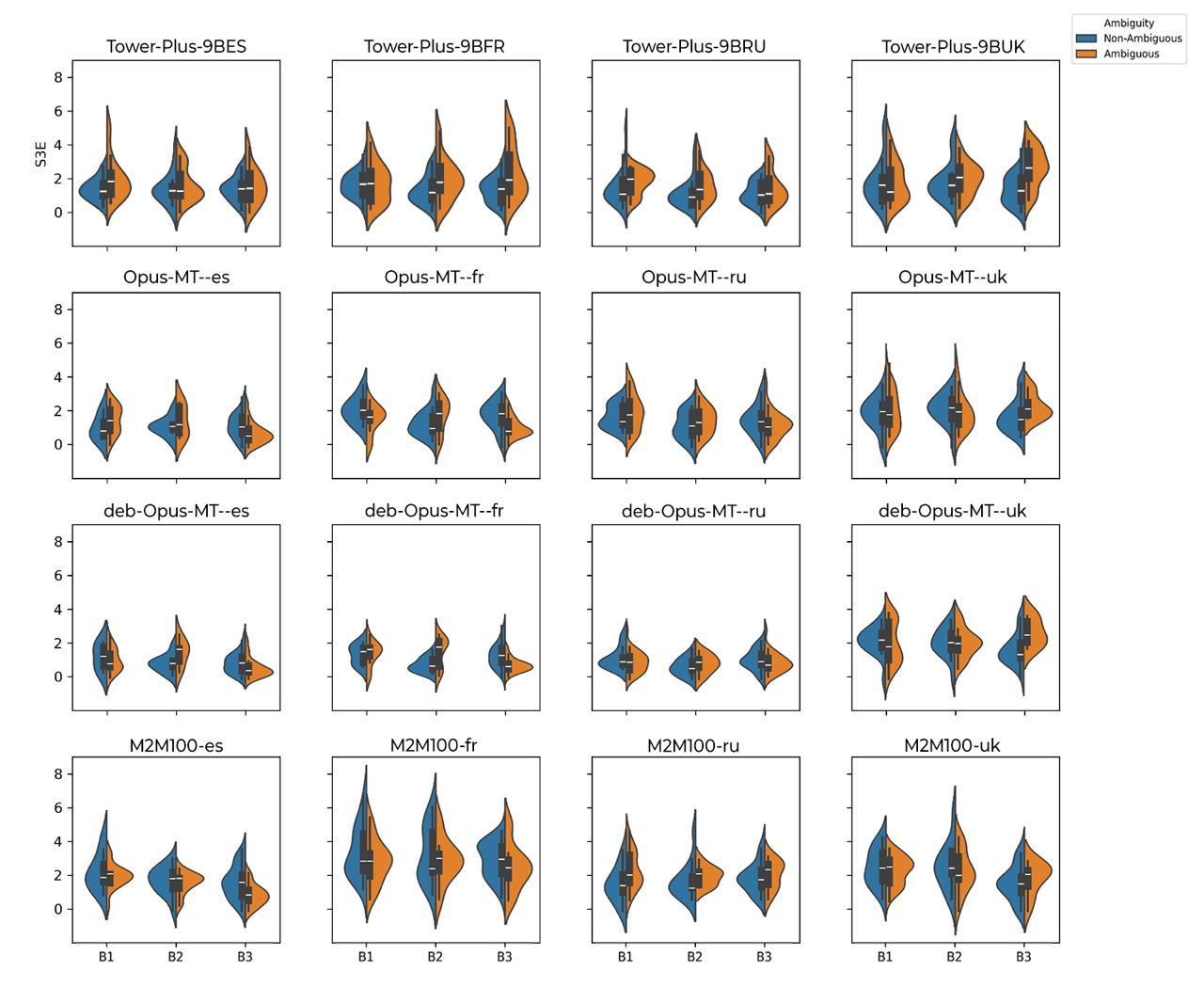}
    \caption{Violin plots of $\mathcal{H}$ (\textsc{s3e}) on ambiguous and unambiguous \textsc{WinoMT} instances. Binned by low (B1), medium (B2) and high (B3) \textsc{comet} scores against human translations.} 
    \label{fig:entropy_vs_comet}
\end{figure*}

In Figure~\ref{fig:entropy_vs_comet}, results on \textsc{WinoMT} are grouped by \textsc{comet} score bins for a more fine-grained analysis at the instance level. 
For the models with negative $\Delta\mathcal{H}$ (\textsc{s3e}) scores, $\Delta \mathcal{H}$ is typically most pronounced for the highest-accuracy translations (e.g. ambiguous scores for \textsc{Tower+9B}-UK in Bin 3 are substantially higher than B1 and B2).
Although debiasing does not reduce the overall $\Delta\mathcal{H}$ score for Ukrainian (see Table~\ref{tab:s3e_entropy_structured_sorted}), it results in the largest improvement in the highest-quality translations: in bin B3, ambiguous $\mathcal{H}$ scores for deb-\textsc{Opus-MT}-UK are notably higher than those of the original model. This improvement is further supported by a substantial 8.41\% drop in masculine focus noun inflections for Ukrainian, compared to 0.88--2.49\% for other languages.  
We hypothesise that this is due to the limited training data in Ukrainian, which may lead to a less stable model that performs worse overall but is more responsive to debiasing in high-quality outputs.
The relationship between translation accuracy and bias under ambiguity appears to differ depending on the level of analysis:  
\textit{between} models, higher accuracy does \textit{not} imply lower bias on ambiguous instances, whereas \textit{within} models, higher-accuracy instances tend to show \textit{lower} bias.

\paragraph{Qualitative Analysis}
\label{sub:qualitative}

A qualitative analysis of the example in Table~\ref{tab:winomt}, presented with corresponding $\mathcal{H}$ values in Table~\ref{tab:winomt_entropy}, corroborates the quantitative findings in Figure~\ref{fig:entropy_vs_comet}.
When translating the sentence in row 1, across all four target languages used in \textsc{WinoMT}, \textsc{Opus-MT} models consistently produce only the masculine variants of the focus noun (`\textit{El mecánico}', `\textit{Le mécanicien}', `\textit{\foreignlanguage{russian}{Механик}}' and `\textit{\foreignlanguage{russian}{Механік}}').
In the anti-stereotypical case (row 2), all languages except Russian include both masculine and feminine forms (`\textit{La mecánica}', `\textit{La mécanicienne}' and `\textit{\foreignlanguage{russian}{Механіка}}'), indicating that these models are sensitive to the masculine stereotype even when the referent in the context is clearly feminine. 
For Russian, the models fail to generate any feminine constructions, even when the context is unambiguously feminine.  
This difference is evident in the $\mathcal{H}$ (\textsc{s3e}) scores, which are higher in row 2 than row 1 for the first three languages. 
Moreover, in the ambiguous case (row 3), all \textsc{Opus-MT} models produce only masculine nouns, regardless of language. 
Consequently, $\mathcal{H}$ (\textsc{s3e}) scores are generally higher for the unambiguous cases (mean of rows 1 and 2) than for the ambiguous case (row 3), except for Russian, where $\mathcal H$ remains low across all conditions.  
These observations are consistent with expectations for biased models: they default to stereotypical gender realisations when the pronoun is ambiguous and sometimes even when the context clearly suggests an anti-stereotypical interpretation.

\begin{table*}[ht]
\centering
\footnotesize
\renewcommand{\arraystretch}{1}
\begin{adjustbox}{max width=\textwidth}
\begin{tabular}{l|cc|cc|cc|cc}
\toprule
\textbf{Sentence} &
\multicolumn{2}{c|}{\textbf{ES} $\rightarrow$ deb} &
\multicolumn{2}{c|}{\textbf{FR} $\rightarrow$ deb} &
\multicolumn{2}{c|}{\textbf{UK} $\rightarrow$ deb} &
\multicolumn{2}{c}{\textbf{RU} $\rightarrow$ deb} \\
\midrule
The \textbf{mechanic} called to inform someone that \textit{he} had completed the repair.
& 0.75 & 0.82
& 0.00 & 0.00
& 2.41 & 1.57
& 0.33 & 0.31 \\
The \textbf{mechanic} called to inform someone that \textit{she} had completed the repair.
& 1.64 & 1.85
& 0.53 & 0.56
& 2.75 & 2.05
& 0.00 & 0.00 \\
The \textbf{mechanic} called to inform someone that \textit{they} had completed the repair.
& 0.74 & 0.87
& 0.00 & 0.00
& 2.38 & 2.03
& 0.37 & 0.37 \\
\bottomrule
\end{tabular}
\end{adjustbox}
\caption{\textsc{WinoMT} examples with $\mathcal{H}$ (\textsc{s3e}) values, for \textsc{Opus-MT} (left) and deb-\textsc{Opus-MT} (right) models.}
\label{tab:winomt_entropy}
\end{table*}

The qualitative analysis also reveals interesting effects of debiasing.
In the anti-stereotypical case in row 2, debiasing increases the number of feminine constructions generated in Spanish (from 43/128 to 55/128, corresponding to a slight increase in $\mathcal{H}$, as feminine forms remain a minority) and Ukrainian (from 72/128 to 128/128, reflected in a decrease in $\mathcal{H}$).  
No notable changes are observed for French or Russian, consistent with stable $\mathcal{H}$ (\textsc{s3e}) scores. 
When it comes to the ambiguous pronoun (row 3), the debiased models continue to generate only masculine variants of `\textit{mechanic}' in Spanish, French and Russian, with $\mathcal{H}$ remaining largely unchanged.  
In contrast, all debiased model outputs in Ukrainian include a feminine translation of the noun, corresponding to a decrease in $\mathcal{H}$ from \textsc{Opus-MT} to debiased \textsc{Opus-MT} in row 3.  
This pattern illustrates that when debiasing leads to overgeneration of feminine morphology in ambiguous contexts, our proposed metric flags this as increased bias (positive $\Delta \mathcal{H}$), indicating that such changes are not deemed as improvements.

\section{Conclusion}
\label{sec:conclusion}

In this work, we apply distribution-level UQ to evaluate bias in MT models.
This method complements gender accuracy, particularly where gender accuracy is inapplicable.
Specifically, it captures the more subtle manifestations of gender bias that arise when models show a preference for one gender in ambiguous contexts. 
Our overall contribution is the novel use of UQ as a bias metric in MT, which 1) does not rely on gender references, 2) is general and captures multiple types of bias, 3) is validated by the established metric of gender accuracy, and 4) provides new insights into biased behavior in ambiguous contexts, a setting not previously studied. 
Future work will extend the proposed bias evaluation method to tasks beyond translation. 

\section*{Limitations}

This study is limited to Romance and Slavic languages, not including many other language families which mark gender and express stereotypes in diverse ways. 
While we tried to account for language differences by including different names for different target languages, accounting for specific masculine-only nouns in Russian, debiasing with language-specific vocabularies, etc., some linguistic idiosyncracies are still not accounted for, such as the fact that profession stereotypes are defined in English and may apply differently in different regions.
Finally, our work is limited to two grammatical genders, and treats `they' as a neutral pronoun that may refer to any gender, however we do not study the interpretation of the pronoun as referring to non-binary people specifically. 
Further directions include applying UQ to ambiguity detection, which could enable more gender-inclusive translations through morphological doubling, where both masculine and feminine morphemes are included for gender neutrality. 
Future work should address these directions. 

\section*{Ethics Statement}

The models used in this study, like all ML models, can be biased and make mistakes, including in gender attribution. 
Our contribution aims to specifically tackle masculine and feminine gender stereotypes via more stringent evaluation metrics, in order to avoid the perpetuation of gender bias. 


\bibliography{custom}

\appendix

\section{Effects of Different Biases}
\label{app:biases}

\begin{table*}[bh]
    \centering
    \footnotesize
    \renewcommand{\arraystretch}{0.95}
    \setlength{\tabcolsep}{4pt}
    \begin{tabular}{l|c|c|c|c|c}
        \hline
        Sentence &
        {Pronoun} &
        {Stereotype} &
        {Subject} &
        {Recency} &
        {IC} \\
        \hline
        The \textbf{mechanic} called to inform someone that \textit{he} had completed the repair. & M & M & M & N & N \\
        The \textbf{mechanic} called to inform someone that \textit{she} had completed the repair. & F & M & F & N & N \\
        The \textbf{mechanic} called to inform someone that \textit{they} had completed the repair. & N & M & N & N & N \\
        \hline
    \end{tabular}
    \caption{\textsc{WinoMT}~\cite{stanovsky2019evaluating} examples with additional annotations of bias cues.}
    \label{tab:winomt2}
\end{table*}

This dataset also contains other linguistic phenomena that were not  explicitly annotated in its released version. Thus, 
we automatically annotate additional linguistic bias cues, namely subject, recency, implicit causality, and person names, using syntactic parses with Spacy(\url{https://spacy.io/}). We release the additional annotations to the public for reproducibility. 

For the Subject bias, following the literature on human biases for coherence \citep{NIEUWLAND2006155}, we hypothesise that models may assume coreference between the subject of the main clause (often the focus noun) and the subject of the complement clause (often the pronoun).
In the example in Table~\ref{tab:winomt2} `\textit{the mechanic}' is the subject, therefore the Subject bias primes an interpretation in which the gender of the subject aligns with that of the subsequent pronoun (M in row 1, F in row 2). 
Furthermore, person names have been shown to have a strong effect on pronoun resolution \citep{saunders2023gender}.
To assess the impact of person names on gender translation, we augment the dataset with common feminine and masculine names matching the gender of the pronoun, selected for their cross-linguistic familiarity (see Appendix~\ref{app:names}).
As an example, when translating into French, the second sentence from Table~\ref{tab:winomt2} would read ``\textit{The mechanic \underline{Anne} called to inform someone that she had completed the repair.}"

Recency bias elicits the attribution of the gender of the most recent noun phrase to the following pronoun \citep{gautam2024robust}. However, in our example, the most recent noun phrase is `\textit{someone}', so the focus noun is unaffected. 
The implicit causality (IC) bias induces an expectation in humans and LMs that when an IC verb is followed by a causal connective, the following pronoun will corefer with either the Subject or the Object of the IC verb, depending on the causal inference \citep{kementchedjhieva2021john}. 
For instance, a clause with the verb `\textit{call}', when followed by an explanation starting with `\textit{because}', is expected to be followed by a mention of the \textit{caller} rather than the \textit{callee}. 
We use an IC verb corpus \citep{garnham2021implicit} for the annotation and find that about 10\% of the dataset contains IC structures. 
The sentences in Table~\ref{tab:winomt2} do not include a causal complement, hence the IC bias does not apply. 

To evaluate model bias,
we assess how bias cues influence the diversity of gender markers in translations. Specifically, we perform an analysis of variance (\textsc{anova}) to examine the effect of bias cues (independent variables) on normalised entropy measures norm-$\mathcal{H}$ (\textsc{s3e}), norm-$\mathcal{H}$ (\textsc{se}), and norm-$\mathcal{H}$ (\textsc{ge}) (dependent variables), using T-tests for significance. This analysis incorporates linguistic biases not previously explored in \textsc{WinoMT}.

Our expectation is that the
entropy of unbiased models should not be affected by the presence of bias cues.
Thus,
we define \textit{normalised entropy}, which compares the $\mathcal{H}$ of a source sentence \( x \) to the average entropy across its \textit{contrast set} \( \mathcal{G}_x \). \( \mathcal{G}_x \) is a group of minimally different sentences that are identical to \( x \) apart from the pronoun (e.g., `\textit{she}', `\textit{he}', `\textit{they}'), including \( x \) itself. The three sentences in Table~\ref{tab:winomt} comprise a \( \mathcal{G}_x \). Formally:
\begin{equation}
\text{norm-}\mathcal{H}(x) = \frac{\mathcal{H}(\mathcal{Y}_x)}{\frac{1}{|\mathcal{G}_x|} \sum_{x' \in \mathcal{G}_x} \mathcal{H}(\mathcal{Y}_{x'})},
\label{eq:normentropy}
\end{equation}

\noindent 
This formulation isolates variation in $\mathcal{H}(\mathcal{Y}_{x})$ attributable specifically to gender by holding all other lexical, syntactic, and semantic content constant across the contrast set.

The results of the experiment presented in Table~\ref{tab:anova} show that most bias cues in the data have a significant effect on the variance of norm-$\mathcal{H}$ (\textsc{s3e}), indicating that the tested models exhibit various social and linguistic biases.\footnote{
norm-$\mathcal{H}$ (\textsc{s3e}) shows the strongest sensitivity to bias cues compared to norm-$\mathcal{H}$ (\textsc{se}) and norm-$\mathcal{H}$ (\textsc{ge}). 
We also experiment with unnormalised $\mathcal{H}$ scores, the results of which are less comparable across metrics, bias types, and models.
The full results are presented in Appendix~\ref{app:anova}.
All trends observed for norm-$\mathcal{H}$ (\textsc{s3e}) are also present in norm-$\mathcal{H}$ (\textsc{se}) and norm-$\mathcal{H}$ (\textsc{ge}), as well as for unnormalised $\mathcal{H}$.}
The results corroborate previous findings. 
The high absolute coefficient values in the Names column indicate that person names have an effect on gender translation even when a disambiguating pronoun is present. 
This is in line with \citet{saunders2023gender}, who have shown that both pronouns and names induce gender bias and are often not sufficient for full disambiguation. 
Secondly, the fact that some Russian nouns have a default masculine grammatical gender regardless of the context \citep{komova2024gender} is reflected in significant decrease in gender diversity for sentences containing such nouns (negative coefficients in the Default M columns indicating lower norm-$\mathcal{H}$). 
Thirdly, we observe that
masculine biases generally reduce norm-$\mathcal H$ (negative coefficients in the M columns), while feminine biases tend to increase it (positive coefficients in the F columns).
This suggests a general default toward masculine translations in models, with outputs becoming more similar under masculine biases and more varied under feminine ones.
This finding aligns with \citet{kuzucu2024uncertainty}, who show that model uncertainty is typically higher for minority groups.
The results in the Ambiguity column in Table~\ref{tab:anova} indicate that norm-$\mathcal H$ increases for Ukrainian and Russian (positive coefficients), but not for Spanish and French, which mirrors the results from Section~\ref{sec:results}. Higher norm-$\mathcal H$ for Ambiguous items (or negative $\Delta\mathcal{H}$) is expected for an unbiased model.

\begin{table*}[ht]
\centering
\scriptsize
\setlength{\tabcolsep}{2pt}
\begin{tabular}{ll|c|cc|cccc|cccc|cc|cccc|cc|c}
\toprule
\multirow{2}{*}{Lang.} & \multirow{2}{*}{Model} & \multirow{2}{*}{Names} & 
\multicolumn{2}{c|}{Recency} & 
\multicolumn{4}{c|}{Implicit Causality} & 
\multicolumn{4}{c|}{Stereotype} & 
\multicolumn{2}{c|}{Subject} &
\multicolumn{4}{c|}{Pronoun} &
\multicolumn{2}{c|}{Default M}&
\multirow{2}{*}{Ambiguity}\\
\cmidrule(lr){4-5}
\cmidrule(lr){6-9}
\cmidrule(lr){10-13}
\cmidrule(lr){14-15}
\cmidrule(lr){16-19}
\cmidrule(lr){20-21}
& & & F & M & S F & S M & O F & O M & S F & S M & O F & O M & F & M & S F& S M& O F& O M& S & O&\\
\midrule
\multirow{3}{*}{es}&\textsc{Opus-MT}&\textbf{0.41} & \textbf{0.41} & -0.05  & \textbf{0.25} & \textbf{-0.19} & \textbf{0.24} & \textbf{-0.33} & 0.06 & 0.10 & \textbf{0.13} & \textbf{0.17} & \textbf{0.38} & \textbf{-0.21} & \textbf{0.24} & \textbf{-0.31} & \textbf{0.29} & \textbf{-0.13}& N/A & N/A& \textbf{-0.18}\\
&deb-\textsc{Opus-MT}&\textbf{-0.05} & \textbf{0.30} & \textbf{-0.11}  & \textbf{0.27} & \textbf{-0.20} & \textbf{0.16} & \textbf{-0.38} & 0.05 & \textbf{0.14} & \textbf{0.08} & 0.05 & \textbf{0.49} & \textbf{-0.14} & \textbf{0.39} & \textbf{-0.24} & \textbf{0.14} & \textbf{-0.21}& N/A & N/A& \textbf{-0.10}\\
&\textsc{m2m100}&\textbf{0.14} & \textbf{0.33}  & \textbf{-0.11} & \textbf{0.29} & \textbf{-0.42} & \textbf{0.28} & \textbf{-0.25} & \textbf{0.18} & \textbf{0.27} & \textbf{0.12} & 0.07 & \textbf{0.51} & -0.04 & \textbf{0.52} & \textbf{-0.08} & 0.04 & \textbf{-0.35}& N/A & N/A& \textbf{-0.11}\\
\midrule
\multirow{3}{*}{fr}&\textsc{Opus-MT}&\textbf{0.54} & \textbf{0.42} & \textbf{0.16}  & 0.05 & \textbf{-0.34} & 0.06 & \textbf{-0.24} & \textbf{0.43} & \textbf{0.45} & \textbf{0.26} & \textbf{0.21} & \textbf{0.16} & \textbf{-0.14} & \textbf{0.16} & \textbf{-0.17} & \textbf{0.20} & -0.04& N/A & N/A& \textbf{-0.29}\\
&deb-\textsc{Opus-MT}&\textbf{0.19} & \textbf{0.22} & 0.02  & 0.05 & \textbf{-0.25} & \textbf{0.17} & \textbf{-0.11} & \textbf{0.23} & \textbf{0.32} & \textbf{0.23} & \textbf{0.13} & \textbf{0.24} & -0.03 & \textbf{0.31} & \textbf{-0.08} & 0.01 & \textbf{-0.14}& N/A & N/A& \textbf{-0.12}\\
&\textsc{m2m100}&\textbf{-0.12} & 0.11 & \textbf{-0.23}  & \textbf{0.50} & 0.09 & \textbf{0.49} & 0.03 & -0.03 & 0.11 & \textbf{0.21} & 0.08 & \textbf{0.70} & \textbf{0.31} & \textbf{0.47} & 0.05 & \textbf{-0.08} & \textbf{-0.40}& N/A & N/A& 0.06\\
\midrule
\multirow{3}{*}{uk}&\textsc{Opus-MT}&\textbf{-0.27} & 0.00 & \textbf{-0.11}  & \textbf{0.26} & -0.02 & \textbf{0.19} & \textbf{0.13} & -0.11 & \textbf{0.17} & 0.03 & \textbf{-0.14} & \textbf{0.27} & \textbf{0.21} & \textbf{0.22} & \textbf{0.13} & \textbf{-0.11} & \textbf{-0.23} &  N/A & N/A& 0.06\\
&deb-\textsc{Opus-MT}&\textbf{-0.43} & -0.06 & \textbf{-0.12} & \textbf{0.41} & 0.00& \textbf{0.18} & 0.07 & \textbf{-0.24} & 0.01 & 0.01 & \textbf{-0.17} & \textbf{0.23} & \textbf{0.17} & \textbf{0.16} & \textbf{0.10} & \textbf{-0.11} & \textbf{-0.17} &  N/A & N/A & \textbf{0.09}\\
&\textsc{m2m100}&\textbf{0.08} & -0.04 & \textbf{-0.18}  & \textbf{0.27} & 0.02 & \textbf{0.33} & \textbf{0.18} & \textbf{0.19} & \textbf{0.37} & -0.03 & \textbf{-0.17} & \textbf{0.53} & \textbf{0.34} & \textbf{0.50} & \textbf{0.26} & \textbf{-0.30} & \textbf{-0.42} & N/A & N/A& \textbf{0.11}\\
\midrule
\multirow{3}{*}{ru}&\textsc{Opus-MT}&0.01 & \textbf{-0.28} & \textbf{-0.41} & 0.00& -0.12 & 0.08 & \textbf{-0.10} & \textbf{-0.19} & \textbf{-0.20} & \textbf{-0.41} & \textbf{-0.39} & \textbf{0.07} & -0.03 & \textbf{0.14} & 0.03 & \textbf{-0.10} & \textbf{-0.24} & \textbf{-0.40} & 0.04& \textbf{0.35} \\
&deb-\textsc{Opus-MT}&\textbf{0.23} & \textbf{-0.10} & \textbf{-0.17}  & 0.05 & 0.03 & 0.05 & -0.03 & \textbf{-0.10} & -0.04 & \textbf{-0.11} & \textbf{-0.14} & \textbf{0.09} & \textbf{0.04} & \textbf{-0.13} & 0.03 & \textbf{-0.08} & \textbf{-0.13} & \textbf{-0.26} & 0.00& \textbf{0.13}\\
&\textsc{m2m100}&\textbf{-0.74} & \textbf{-0.16} & \textbf{-0.21} & 0.10 & -0.04 & \textbf{-0.12} & \textbf{-0.25} & \textbf{0.12} & \textbf{0.12} & \textbf{-0.20} & \textbf{-0.18} & \textbf{0.06} & -0.02 & \textbf{0.08} & 0.01 & \textbf{-0.07} & \textbf{-0.11} & \textbf{-0.20} & \textbf{-0.22} & \textbf{0.18}\\
\bottomrule
\end{tabular}
\caption{\textsc{anova} results: single effects of bias cues (\textbf{F}eminine, \textbf{M}asculine, \textbf{S}ubject and \textbf{O}bject) on norm-$\mathcal{H}$ (\textsc{s3e}). Values correspond to effect coefficients (deviations from a reference group). \textbf{Boldface} indicates statistical significance (p < 0.05). The sign of the values indicates whether the presence of the variable increases (positive) or decreases (negative) the mean $\mathcal H$ of the group containing the given variable value. Reference group is N for all columns except: `no name' for Names, `no default' for Default M, `unambiguous' for Ambiguity.}
\label{tab:anova}
\end{table*}

Table~\ref{tab:anova} further demonstrates the impact of debiasing, with effect sizes typically reduced (smaller absolute coefficients) in debiased models versus non-debiased ones across languages, supporting the partial effectiveness of debiasing.

\section{Further UQ discussion}
\label{app:alpha}

\citet{farquhar2024detecting} does not define a per-element surprisal; the original definition computes Shannon entropy over clusters. $\mathcal{Y}$ are mapped to clusters $\mathcal{C}$, and \text{SE} is:

\[
\mathcal{H}_{\text{SE}}(\mathcal{C})=-\underset{c\sim\mathcal{C}}{\mathbb{E}} \log p(c|x).
\]
Surprisal is thus defined for a cluster instead of an element, but it is easy to show that our per-element surprisal obtains equivalent entropy as the original definition.

\citet{cheng2024measuring} introduce a hyperparameter $\alpha$ which is applied as an exponent to the similarity function. This is used to rescale $\mathcal{S}$ for more favorable performance on benchmarks. We tune $\alpha$ for the highest correlation between \textsc{s3e} and the entropy of the gender labels assigned to the nouns in question by the morphological parser. 
This way we aim for $\mathcal H$ (\textsc{s3e}) to capture gender variation, with higher values of $\mathcal H$ (\textsc{s3e}) indicating more diversity in the gender morphemes. 

We also experiment with alternative similarity metrics to cosine similarity, such as Euclidean Distance, Chebyshev Distance, Manhattan Distance, Minkowski Distance, etc. but find no notable differences and use the simplest option of Cosine Similarity in the paper. 

\section{Annotation of Names}
\label{app:names}

Table~\ref{tab:names} presents the names used for expanding the \textsc{WinoMT} dataset to include commonly used person names for masculine and feminine genders in French, Spanish, Ukrainian and Russian. 

\begin{table}[ht]
    \centering
    \footnotesize{
    \begin{tabularx}{\linewidth}{|X|X|X|X|X|}
        \hline
         & Spanish & French & Ukrainian & Russian \\
        \hline
        F & Carla, Lucía, María, Sofía, Ana & Anne, Camille, Élise, Sophie, Claire & Anna, Olena, Sofiia, Natalka, Oksana & Anna, Maria, Elena, Svetlana, Natalia \\
        \hline
        M & Gabriel, Carlos, Luis, Javier, Miguel & Victor, Julien, Antoine, Lucas, Mathieu & Ivan, Andrii, Taras, Oleh, Mykola & Ivan, Dmitry, Alexei, Sergey, Nikolai \\
        \hline
    \end{tabularx}}
    \caption{The names used for extending the \textsc{WinoMT} dataset for different target languages.}
    \label{tab:names}
\end{table}

\section{Translation Guidelines and Translator Recruitment Details}
\label{app:translationguidelines}

Professional translators were hired and paid £0.2 per word as requested. 
The translation guidelines are as follows: 

\textit{Please translate the following sentences into [French\slash Spanish\slash Ukrainian\slash Russian]. Please provide two translations for each sentence, with different gender inflections for the highlighted noun —one masculine and one feminine. Please keep the pronoun true to the original to the extent that it is possible. Please use the translations that appear natural to you as a native speaker rather than using forms that are prescribed\footnote{This instruction was added due to inquiries by translators about the use of formal or dominant dialects as well as language policy changes influencing how they might translate some words.}. Please note whether both translations are acceptable in terms of the gender of the noun matching the contextual information, and if they are not, which one is the correct one:}

\begin{itemize}
    \item The \textbf{developer} argued with the designer because she did not like the design.
    \item The undergraduate presented the \textbf{administrator} with a petition that they had organized.
\end{itemize}

\section{Human Annotation Quality}
\label{app:cohen}

The annotations of Correct and Incorrect sentences by professional translators are compared to the gold standard gender annotations in the dataset, and across all four translators range between 93.17 and 93.27 in Cohen's $\kappa$ scores, confirming that apart from some linguistic idiosyncrasies of each language (e.g. `victim' in Spanish is always feminine and so regardless of the contextualising pronoun will take the same form), the annotators agree on which sentences should be correctly translated in which gender.

\section{Overall Model Performance}
\label{app:models}

Table~\ref{tab:models} presents the performance of the models used in this study in terms of the \textsc{comet} metric\footnote{\url{https://huggingface.co/Unbabel/wmt22-comet-da}} \citep{rei-etal-2022-comet} on WMT datasets which contain the target languages \citep{ws-2012-statistical, ws-2013-statistical, ws-2014-statistical, wmt-2023, wmt-2024-1}. 
The models are run on a single NVIDIA TU102 GPU.

\begin{table*}[ht]
    \centering
    \footnotesize{
    \begin{tabular}{l l | c c c c c c c}
        \hline
        Dataset & Model & es & fr & uk & ru & de & it & el \\
        \hline
        newstest2012 & Opus-MT        & 84.52 & 82.21 & --    & --    & -- & -- & -- \\
                     & deb-Opus-MT    & 84.47 & 82.22 & --    & --    & -- & -- & -- \\
                     & m2m100         & 84.23 & 82.69 & --    & --    & -- & -- & -- \\
                     & Tower+9B       & 86.61 & 84.86 & --    & --    & -- & -- & -- \\
        \hline
        newstest2013 & Opus-MT        & 85.28 & 83.45 & --    & --    & -- & -- & -- \\
                     & deb-Opus-MT    & 85.24 & 83.44 & --    & --    & -- & -- & -- \\
                     & m2m100         & 84.83 & 83.66 & --    & --    & -- & -- & -- \\
                     & Tower+9B       & 87.14 & 86.27 & --    & --    & -- & -- & -- \\
        \hline
        newstest2014 & Opus-MT        & --    & 85.01 & --    & 87.44 & -- & -- & -- \\
                     & deb-Opus-MT    & --    & 85.00 & --    & 87.44 & -- & -- & -- \\
                     & m2m100         & --    & 85.28 & --    & 71.95 & -- & -- & -- \\
                     & Tower+9B       & --    & 88.71 & --    & 93.04 & -- & -- & -- \\
        \hline
        wmttest2023  & Opus-MT        & --    & --    & 74.58 & 79.02 & -- & -- & -- \\
                     & deb-Opus-MT    & --    & --    & 74.58 & 79.02 & -- & -- & -- \\
                     & m2m100         & --    & --    & 73.58 & 80.23 & -- & -- & -- \\
                     & Tower+9B       & --    & --    & 86.24 & 83.75 & -- & -- & -- \\
        \hline
        wmttest2024  & Opus-MT        & --    & --    & 66.99 & 71.64 & -- & -- & -- \\
                     & deb-Opus-MT    & --    & --    & 67.00 & 71.63 & -- & -- & -- \\
                     & m2m100         & --    & --    & 79.45 & 89.30 & -- & -- & -- \\
                     & Tower+9B       & --    & --    & 87.07 & 85.69 & -- & -- & -- \\
        \midrule
        newstest2009  & Opus-MT        & --    & --    & --    & --    & 81.30 & 84.65 & -- \\
                     & deb-Opus-MT    & --    & --    & --    & --    & 81.24 & 84.58 & -- \\
                     & m2m100         & --    & --    & --    & --    & 82.29 & 85.04 & -- \\
                     & Tower+9B       & --    & --    & --    & --    & 85.33 & 87.78 & -- \\
        \midrule
        newstest2024  & Opus-MT        & --    & --    & --    & --    & -- & -- & 78.41 \\
                     & deb-Opus-MT    & --    & --    & --    & --    & -- & -- & 77.90 \\
                     & m2m100         & --    & --    & --    & --    & -- & -- & 78.58 \\
                     & Tower+9B       & --    & --    & --    & --    & -- & -- &  66.23\\
        \midrule
        mean         & Opus-MT        & 84.90 & 83.56 & 70.79 & 79.37 & 81.30 & 84.65 & 78.41 \\
                     & deb-Opus-MT    & 84.86 & 83.55 & 70.79 & 79.36 & 81.24 & 84.58 & 77.90 \\
                     & m2m100         & 84.53 & 83.88 & 76.52 & 80.49 & 82.29 & 85.04 & 78.58 \\
                     & Tower+9B       & 86.88 & 86.61 & 86.66 & 87.49 & 85.33 & 87.78 &  66.23\\
        \hline
    \end{tabular}}
    \caption{\textsc{comet} scores on WMT test sets for the models used.}
    \label{tab:models}
\end{table*}

\clearpage 
\section{\textsc{anova} Results}
\label{app:anova}

Table~\ref{tab:fullanova} presents the \textsc{anova} results for \textsc{s3e}, \textsc{se} and \textsc{ge} metrics. 
Table~\ref{tab:fullanova_nonorm} presents the \textsc{anova} results without normalising the $\mathcal{H}$ values. 

\begin{table*}[ht]
\centering
\scriptsize
\setlength{\tabcolsep}{2pt}
\begin{tabular}{ll|c|cc|cccc|cccc|cc|cccc|cc|c}
\toprule
\multirow{2}{*}{Lang.} & \multirow{2}{*}{Model} & \multirow{2}{*}{Names} & 
\multicolumn{2}{c|}{Recency} & 
\multicolumn{4}{c|}{Implicit Causality} & 
\multicolumn{4}{c|}{Stereotype} & 
\multicolumn{2}{c|}{Subject} &
\multicolumn{4}{c|}{Context} &
\multicolumn{2}{c|}{Default M}&
\multirow{2}{*}{Ambiguity}\\
\cmidrule(lr){4-5}
\cmidrule(lr){6-9}
\cmidrule(lr){10-13}
\cmidrule(lr){14-15}
\cmidrule(lr){16-19}
\cmidrule(lr){20-21}
& & & F & M & S F & S M & O F & O M & S F & S M & O F & O M & F & M & S F& S M& O F& O M& S & O&\\
\midrule
\multicolumn{22}{c}{\textsc{s3e}}\\
\midrule
\multirow{3}{*}{es}&\textsc{Opus-MT}&\textbf{0.41} & \textbf{0.41} & -0.05  & \textbf{0.25} & \textbf{-0.19} & \textbf{0.24} & \textbf{-0.33} & 0.06 & 0.10 & \textbf{0.13} & \textbf{0.17} & \textbf{0.38} & \textbf{-0.21} & \textbf{0.24} & \textbf{-0.31} & \textbf{0.29} & \textbf{-0.13}& N/A & N/A& \textbf{-0.18}\\
&deb-\textsc{Opus-MT}&\textbf{-0.05} & \textbf{0.30} & \textbf{-0.11}  & \textbf{0.27} & \textbf{-0.20} & \textbf{0.16} & \textbf{-0.38} & 0.05 & \textbf{0.14} & \textbf{0.08} & 0.05 & \textbf{0.49} & \textbf{-0.14} & \textbf{0.39} & \textbf{-0.24} & \textbf{0.14} & \textbf{-0.21}& N/A & N/A& \textbf{-0.10}\\
&\textsc{m2m100}&\textbf{0.14} & \textbf{0.33}  & \textbf{-0.11} & \textbf{0.29} & \textbf{-0.42} & \textbf{0.28} & \textbf{-0.25} & \textbf{0.18} & \textbf{0.27} & \textbf{0.12} & 0.07 & \textbf{0.51} & -0.04 & \textbf{0.52} & \textbf{-0.08} & 0.04 & \textbf{-0.35}& N/A & N/A& \textbf{-0.11}\\
\multirow{3}{*}{fr}&\textsc{Opus-MT}&\textbf{0.54} & \textbf{0.42} & \textbf{0.16}  & 0.05 & \textbf{-0.34} & 0.06 & \textbf{-0.24} & \textbf{0.43} & \textbf{0.45} & \textbf{0.26} & \textbf{0.21} & \textbf{0.16} & \textbf{-0.14} & \textbf{0.16} & \textbf{-0.17} & \textbf{0.20} & -0.04& N/A & N/A& \textbf{-0.29}\\
&deb-\textsc{Opus-MT}&\textbf{0.19} & \textbf{0.22} & 0.02  & 0.05 & \textbf{-0.25} & \textbf{0.17} & \textbf{-0.11} & \textbf{0.23} & \textbf{0.32} & \textbf{0.23} & \textbf{0.13} & \textbf{0.24} & -0.03 & \textbf{0.31} & \textbf{-0.08} & 0.01 & \textbf{-0.14}& N/A & N/A& \textbf{-0.12}\\
&\textsc{m2m100}&\textbf{-0.12} & 0.11 & \textbf{-0.23}  & \textbf{0.50} & 0.09 & \textbf{0.49} & 0.03 & -0.03 & 0.11 & \textbf{0.21} & 0.08 & \textbf{0.70} & \textbf{0.31} & \textbf{0.47} & 0.05 & \textbf{-0.08} & \textbf{-0.40}& N/A & N/A& 0.06\\
\multirow{3}{*}{uk}&\textsc{Opus-MT}&\textbf{-0.27} & 0.00 & \textbf{-0.11}  & \textbf{0.26} & -0.02 & \textbf{0.19} & \textbf{0.13} & -0.11 & \textbf{0.17} & 0.03 & \textbf{-0.14} & \textbf{0.27} & \textbf{0.21} & \textbf{0.22} & \textbf{0.13} & \textbf{-0.11} & \textbf{-0.23} &  N/A & N/A& 0.06\\
&deb-\textsc{Opus-MT}&\textbf{-0.43} & -0.06 & \textbf{-0.12} & \textbf{0.41} & 0.00& \textbf{0.18} & 0.07 & \textbf{-0.24} & 0.01 & 0.01 & \textbf{-0.17} & \textbf{0.23} & \textbf{0.17} & \textbf{0.16} & \textbf{0.10} & \textbf{-0.11} & \textbf{-0.17} &  N/A & N/A & \textbf{0.09}\\
&\textsc{m2m100}&\textbf{0.08} & -0.04 & \textbf{-0.18}  & \textbf{0.27} & 0.02 & \textbf{0.33} & \textbf{0.18} & \textbf{0.19} & \textbf{0.37} & -0.03 & \textbf{-0.17} & \textbf{0.53} & \textbf{0.34} & \textbf{0.50} & \textbf{0.26} & \textbf{-0.30} & \textbf{-0.42} & N/A & N/A& \textbf{0.11}\\
\multirow{3}{*}{ru}&\textsc{Opus-MT}&0.01 & \textbf{-0.28} & \textbf{-0.41} & 0.00& -0.12 & 0.08 & \textbf{-0.10} & \textbf{-0.19} & \textbf{-0.20} & \textbf{-0.41} & \textbf{-0.39} & \textbf{0.07} & -0.03 & \textbf{0.14} & 0.03 & \textbf{-0.10} & \textbf{-0.24} & \textbf{-0.40} & 0.04& \textbf{0.35} \\
&deb-\textsc{Opus-MT}&\textbf{0.23} & \textbf{-0.10} & \textbf{-0.17}  & 0.05 & 0.03 & 0.05 & -0.03 & \textbf{-0.10} & -0.04 & \textbf{-0.11} & \textbf{-0.14} & \textbf{0.09} & \textbf{0.04} & \textbf{-0.13} & 0.03 & \textbf{-0.08} & \textbf{-0.13} & \textbf{-0.26} & 0.00& \textbf{0.13}\\
&\textsc{m2m100}&\textbf{-0.74} & \textbf{-0.16} & \textbf{-0.21} & 0.10 & -0.04 & \textbf{-0.12} & \textbf{-0.25} & \textbf{0.12} & \textbf{0.12} & \textbf{-0.20} & \textbf{-0.18} & \textbf{0.06} & -0.02 & \textbf{0.08} & 0.01 & \textbf{-0.07} & \textbf{-0.11} & \textbf{-0.20} & \textbf{-0.22} & \textbf{0.18}\\
\midrule
\multicolumn{22}{c}{\textsc{se}}\\
\midrule
\multirow{3}{*}{es}&\textsc{Opus-MT}&\textbf{-1.64} & \textbf{0.19} & -0.08  & 0.13 & -0.19 & 0.09 & \textbf{-0.29} & -0.01 & 0.02 & 0.06 & 0.07 & \textbf{0.28} & \textbf{-0.12} & \textbf{0.17} & \textbf{-0.22} & \textbf{0.15} & \textbf{-0.08}& N/A & N/A& -0.05\\
&deb-\textsc{Opus-MT}&\textbf{-0.06} & \textbf{0.27} & \textbf{0.13}  & -0.01 & -0.07 & 0.00& \textbf{-0.19} & 0.06 & \textbf{0.10} & \textbf{0.16} & \textbf{0.12} & -0.01 & \textbf{-0.18} & -0.01 & \textbf{-0.19} & \textbf{0.20} & \textbf{0.08}& N/A & N/A& \textbf{-0.20}\\
&\textsc{m2m100}&\textbf{-1.71} & \textbf{0.23} & -0.04 & 0.20 & \textbf{-0.29} & \textbf{0.17} & \textbf{-0.16} & 0.10& \textbf{0.18} & \textbf{0.15} & 0.08 & \textbf{0.28} & -0.06 & \textbf{0.25} & \textbf{-0.09} & \textbf{0.07} & \textbf{-0.16}& N/A & N/A& -0.10\\
\multirow{3}{*}{fr}&\textsc{Opus-MT}&\textbf{-1.59} & \textbf{0.19} & 0.00 & 0.09 & \textbf{-0.25} & 0.05 & \textbf{-0.17} & \textbf{0.23} & \textbf{0.25} & \textbf{0.15} & \textbf{0.11} & \textbf{0.11} & \textbf{-0.12} & \textbf{0.10} & \textbf{-0.15} & \textbf{0.14} & -0.04& N/A & N/A& -0.09\\
&deb-\textsc{Opus-MT}&-0.01 & \textbf{0.24} & \textbf{0.08}  & 0.11 & \textbf{-0.17} & 0.04 & \textbf{-0.10} & \textbf{0.23} & \textbf{0.20} & \textbf{0.19} & \textbf{0.18} & \textbf{-0.06} & \textbf{-0.22} & 0.01 & \textbf{-0.15} & \textbf{0.18} & 0.03& N/A & N/A& \textbf{-0.16}\\
&\textsc{m2m100}&\textbf{-0.15} & \textbf{0.29} & \textbf{0.16}  & 0.03 & -0.05 & 0.02 & \textbf{-0.14} & \textbf{0.27} & \textbf{0.24} & \textbf{0.27} & \textbf{0.30} & 0.01 & \textbf{-0.14} & 0.01 & \textbf{-0.15} & \textbf{0.19} & \textbf{0.06}& N/A & N/A& \textbf{-0.22}\\
\multirow{3}{*}{uk}&\textsc{Opus-MT}&\textbf{-1.54} & 0.00& \textbf{-0.12}  & \textbf{0.21} & -0.05 & 0.10& 0.02 & -0.09 & 0.05 & 0.01 & -0.08 & \textbf{0.13} & 0.04 & \textbf{0.10} & 0.02 & 0.00& \textbf{-0.14} &  N/A & N/A& 0.06\\
&deb-\textsc{Opus-MT}&\textbf{0.09} & \textbf{0.12} & \textbf{-0.06}  & \textbf{0.13} & -0.07 & \textbf{0.08} & 0.01 & -0.02 & \textbf{0.08} & \textbf{0.15} & \textbf{0.09} & \textbf{0.06} & \textbf{-0.11} & \textbf{-0.07} & \textbf{-0.10} & \textbf{0.13} & \textbf{0.06} & N/A & N/A& \textbf{0.09}\\
&\textsc{m2m100}&\textbf{-1.72} & -0.04 & \textbf{-0.14}  & 0.20 & 0.03 & \textbf{0.15} & 0.08 & 0.05 & \textbf{0.16} & -0.02 & \textbf{-0.11} & \textbf{0.22} & \textbf{0.10} & \textbf{0.18} & \textbf{0.09} & \textbf{-0.09} & \textbf{-0.19} &  N/A & N/A& 0.09\\
\multirow{3}{*}{ru}&\textsc{Opus-MT}&\textbf{-1.50} & \textbf{-0.17} & \textbf{-0.25} & \textbf{0.21} & 0.07 & 0.00& -0.02 & \textbf{-0.13}  & \textbf{-0.18} & \textbf{-0.25} & \textbf{-0.28} & 0.03 & -0.03 & \textbf{0.10} & 0.00 & \textbf{-0.06} & \textbf{-0.14} & \textbf{-0.32} & -0.02& \textbf{0.21}\\
&deb-\textsc{Opus-MT}&\textbf{-0.06}& 0.04 & 0.01 & 0.11 & 0.05 & 0.00 & -0.06 & \textbf{-0.22} & \textbf{-0.13} & \textbf{0.07} & -0.01 & \textbf{-0.11} & \textbf{-0.15} & \textbf{-0.05} & \textbf{-0.11} & \textbf{0.09} & \textbf{0.07} & \textbf{-0.24} & \textbf{-0.12} & 0.03\\
&\textsc{m2m100}&\textbf{-1.57} & -0.09 & \textbf{-0.14} & \textbf{0.11} & 0.10& 0.04 & \textbf{-0.18} & -0.01 & 0.00& \textbf{-0.10} & \textbf{-0.10} & 0.00 & -0.06 & 0.03 & -0.01 & -0.01 & \textbf{-0.07} & \textbf{-0.12} & \textbf{-0.18} & 0.11\\
\midrule
\multicolumn{22}{c}{\textsc{ge}}\\
\midrule
\multirow{3}{*}{es}&\textsc{Opus-MT}&\textbf{0.02} & \textbf{0.30} & \textbf{0.04} & \textbf{-0.17} & \textbf{0.16} & \textbf{-0.10}  & \textbf{-0.14} & \textbf{0.11} & \textbf{0.15} & \textbf{0.17} & \textbf{0.12} & \textbf{0.19} & \textbf{-0.12} & \textbf{0.19} & \textbf{-0.20} & \textbf{0.15} & \textbf{-0.06}& N/A & N/A& \textbf{0.17}\\
&deb-\textsc{Opus-MT}&\textbf{-0.04} & \textbf{0.27} & \textbf{0.06} & \textbf{0.16} & \textbf{-0.10} & \textbf{0.15} & \textbf{-0.12} & \textbf{0.09} & \textbf{0.17} & \textbf{0.17} & \textbf{0.09} & \textbf{0.21} & \textbf{-0.09} & \textbf{0.24} & \textbf{-0.16} & \textbf{0.09} & \textbf{-0.06}& N/A & N/A& \textbf{-0.16} \\
&\textsc{m2m100}&\textbf{-0.10} & \textbf{0.16} & 0.00 & \textbf{0.08} & \textbf{-0.10} & \textbf{0.09} & \textbf{-0.08} & \textbf{0.05} & \textbf{0.07} & \textbf{0.10} & \textbf{0.09} & \textbf{0.09} & \textbf{-0.08} & \textbf{0.08} & \textbf{-0.16} & \textbf{0.12} & -0.01& N/A & N/A& \textbf{-0.08}\\
\multirow{3}{*}{fr}&\textsc{Opus-MT}&0.01 & \textbf{0.20} & \textbf{0.05}  & \textbf{0.12} & \textbf{-0.06} & \textbf{0.07} & \textbf{-0.11} & \textbf{0.09} & \textbf{0.13} & \textbf{0.12} & \textbf{0.08} & \textbf{0.09} & \textbf{-0.10} & \textbf{0.06} & \textbf{-0.19} & \textbf{0.15} & \textbf{0.03}& N/A & N/A& \textbf{-0.13}\\
&deb-\textsc{Opus-MT}&\textbf{-0.05} & \textbf{0.18} & \textbf{0.04} & \textbf{-0.11} & \textbf{0.08}  & \textbf{0.08} & \textbf{-0.11} & \textbf{0.05} & \textbf{0.12} & \textbf{0.12} & \textbf{0.06} & \textbf{0.11} & \textbf{-0.09} & \textbf{0.10} & \textbf{-0.17} & \textbf{0.11} & 0.01& N/A & N/A& \textbf{-0.11}\\
&\textsc{m2m100}&\textbf{-0.02} & \textbf{0.19} & 0.02  & \textbf{0.12} & \textbf{-0.10} & \textbf{0.09} & \textbf{-0.12} & \textbf{0.08} & \textbf{0.10} & \textbf{0.11} & \textbf{0.07} & \textbf{0.08} & \textbf{-0.11} & \textbf{0.04} & \textbf{-0.19} & \textbf{0.17} & \textbf{0.02}& N/A & N/A& \textbf{-0.11}\\
\multirow{3}{*}{uk}&\textsc{Opus-MT}&\textbf{-0.04} & \textbf{0.05} & \textbf{-0.04}  & \textbf{0.07} & -0.03 & 0.03 & \textbf{-0.07} & \textbf{0.06} & \textbf{0.06} & 0.02 & 0.02 & 0.02 & \textbf{-0.06} & \textbf{-0.11} & \textbf{-0.15} & \textbf{0.16} & \textbf{0.06} &  N/A & N/A& 0.01\\
&deb-\textsc{Opus-MT}&\textbf{-0.03} & \textbf{0.05} & 0.02  & \textbf{0.07} & -0.03 & 0.00& \textbf{-0.06} & \textbf{0.06} & \textbf{0.07} & \textbf{0.03} & \textbf{-0.03} & \textbf{-0.02} & \textbf{-0.05} & \textbf{-0.16} & \textbf{-0.17} & \textbf{0.17} & \textbf{0.14} &  N/A & N/A& \textbf{0.03}\\
&\textsc{m2m100}&\textbf{-0.04} & \textbf{0.06} & \textbf{-0.04}  & \textbf{0.09} & -0.05 & \textbf{0.05} & \textbf{-0.06} & \textbf{0.07} & \textbf{0.08} & 0.01 & 0.01 & \textbf{0.03} & \textbf{-0.07} & \textbf{-0.07} & \textbf{-0.18} & \textbf{0.16} & \textbf{0.06} & N/A & N/A& 0.01\\
\multirow{3}{*}{ru}&\textsc{Opus-MT}&-0.01 & 0.00& \textbf{-0.03} & -0.02 & -0.01& 0.00& \textbf{-0.04} & 0.03 & \textbf{0.03} & \textbf{-0.04} & \textbf{-0.06} & -0.01 & \textbf{-0.04} & \textbf{-0.11} & \textbf{-0.13} & \textbf{0.12} & \textbf{0.08} & \textbf{-0.04} & 0.00 & -0.02 \\
&deb-\textsc{Opus-MT}&\textbf{-0.02} & -0.01 & -0.02 & \textbf{-0.05} & \textbf{-0.05} & -0.01 & -0.02 & 0.01 & 0.02 & \textbf{-0.04} & \textbf{-0.05} & \textbf{-0.03} & \textbf{-0.04} & \textbf{-0.15} & \textbf{-0.15} & \textbf{0.13} & \textbf{0.12} & \textbf{-0.05} & -0.01 & -0.02\\
&\textsc{m2m100}&\textbf{0.02} & \textbf{0.07} & 0.02 & -0.01 & \textbf{-0.06} & -0.01 & \textbf{-0.07} & \textbf{0.09} & \textbf{0.07} & 0.02 & \textbf{0.04} & -0.01 & \textbf{-0.06} & \textbf{-0.05} & \textbf{-0.11} & \textbf{0.10} & \textbf{0.06} & \textbf{-0.04} & \textbf{-0.03} & \textbf{0.04}\\
\bottomrule
\end{tabular}
\caption{ 
\textsc{anova} results: single effects of bias cues (\textbf{F}eminine, \textbf{M}asculine, \textbf{S}ubject and \textbf{O}bject) on norm-$\mathcal{H}$ (\textsc{s3e}), norm-$\mathcal{H}$ (\textsc{se}) and norm-$\mathcal{H}$ (\textsc{ge}). Values correspond to effect coefficients (deviations from a reference group). \textbf{Boldface} indicates statistical significance (p < 0.05). The sign of the values indicates whether the presence of the variable increases (positive) or decreases (negative) the mean $\mathcal H$ of the group containing the given variable value. Reference group is N for all columns except: `no name' for Names, `no default' for Default M, `unambiguous' for Ambiguity.
}
\label{tab:fullanova}
\end{table*}

\begin{table*}[ht]
\centering
\scriptsize
\setlength{\tabcolsep}{2pt}
\begin{tabular}{ll|c|cc|cccc|cccc|cc|cccc|cc|c}
\toprule
\multirow{2}{*}{Lang.} & \multirow{2}{*}{Model} & \multirow{2}{*}{Names} & 
\multicolumn{2}{c|}{Recency} & 
\multicolumn{4}{c|}{Implicit Causality} & 
\multicolumn{4}{c|}{Stereotype} & 
\multicolumn{2}{c|}{Subject} &
\multicolumn{4}{c|}{Context} &
\multicolumn{2}{c|}{Default M}&
\multirow{2}{*}{Ambiguity}\\
\cmidrule(lr){4-5}
\cmidrule(lr){6-9}
\cmidrule(lr){10-13}
\cmidrule(lr){14-15}
\cmidrule(lr){16-19}
\cmidrule(lr){20-21}
& & & F & M & S F & S M & O F & O M & S F & S M & O F & O M & F & M & S F& S M& O F& O M& S & O&\\
\midrule
\multicolumn{22}{c}{\textsc{s3e}}\\
\midrule
\multirow{3}{*}{es}&\textsc{Opus-MT}&\textbf{12.12} & \textbf{47.0} & \textbf{83.53} & -2.45 & \textbf{20.67} & -2.44 & \textbf{27.76} & \textbf{38.98} & \textbf{49.12} & \textbf{44.16} & \textbf{37.6} & 0.23 & \textbf{31.36} & \textbf{7.95} & \textbf{42.89} & \textbf{-22.44} & \textbf{14.64} & N/A & N/A& \textbf{-65.29}\\
&deb-\textsc{Opus-MT}&\textbf{-2.35} & \textbf{36.52} & \textbf{79.64} & \textbf{-58.1} & -7.06 & \textbf{18.24} & \textbf{-8.44} & \textbf{31.09} & \textbf{33.94} & \textbf{42.22} & \textbf{36.85} & \textbf{32.48} & -1.86 & \textbf{34.19} & \textbf{6.06} & \textbf{48.72} & \textbf{-29.24} & N/A & N/A& \textbf{14.06}\\
&\textsc{m2m100}&\textbf{0.19} & \textbf{0.54} & 0.02 & \textbf{-0.28} & \textbf{0.38} & \textbf{-0.27} & \textbf{0.37} & \textbf{-0.26} & \textbf{0.18} & \textbf{0.25} & \textbf{0.25} & \textbf{0.17} & \textbf{0.31} & \textbf{-0.25} & \textbf{0.31} & \textbf{-0.29} & \textbf{0.31} & N/A & N/A& \textbf{-0.17}\\
\multirow{3}{*}{fr}&\textsc{Opus-MT}&\textbf{0.12} & \textbf{0.41} & \textbf{0.06} & \textbf{0.32} & \textbf{-0.15} & \textbf{0.23} & \textbf{-0.17} & \textbf{0.18} & \textbf{0.25} & \textbf{0.23} & \textbf{0.14} & \textbf{0.26} & \textbf{-0.16} & \textbf{0.32} & \textbf{-0.19} & \textbf{0.16} & \textbf{-0.14} & N/A & N/A& \textbf{-0.24}\\
&deb-\textsc{Opus-MT}&\textbf{-0.02} & \textbf{0.26} & \textbf{0.06} & \textbf{-0.16} & \textbf{0.23} & \textbf{-0.07} & \textbf{0.15} & \textbf{-0.1} & \textbf{0.12} & \textbf{0.21} & \textbf{0.18} & \textbf{0.08} & \textbf{0.21} & \textbf{-0.08} & \textbf{0.25} & \textbf{-0.12} & \textbf{0.05} & N/A & N/A& \textbf{-0.08}\\
&\textsc{m2m100}&\textbf{-0.12} & 0.11 & \textbf{-0.23} & 0.06 & \textbf{0.5} & 0.09 & \textbf{0.49} & 0.03 & -0.03 & 0.11 & \textbf{0.21} & 0.08 & \textbf{0.7} & \textbf{0.31} & \textbf{0.47} & 0.05 & \textbf{-0.08} & N/A & N/A& \textbf{-0.4}\\
\multirow{3}{*}{uk}&\textsc{Opus-MT}&\textbf{0.12} & \textbf{0.22} & 0.03 & \textbf{0.16} & \textbf{-0.07} & \textbf{0.09} & \textbf{-0.11} & \textbf{0.15} & \textbf{0.18} & \textbf{0.14} & \textbf{0.1} & \textbf{0.09} & \textbf{-0.1} & \textbf{0.05} & \textbf{-0.1} & \textbf{0.16} & \textbf{-0.04} & \textbf{-0.09} & \textbf{-0.05} & \textbf{-0.12}\\
&deb-\textsc{Opus-MT}&0.83 & \textbf{-10.2} & \textbf{24.93} & \textbf{-7.38} & \textbf{-19.32} & \textbf{18.37} & \textbf{-14.82} & \textbf{27.31} & \textbf{5.26} & \textbf{9.67} & \textbf{8.87} & \textbf{5.29} & \textbf{-15.13} & \textbf{23.85} & \textbf{-10.15} & \textbf{44.14} & \textbf{-26.98} & 1.7 & -2.21 & -0.34\\
&\textsc{m2m100}&\textbf{0.14} & \textbf{0.3} & \textbf{0.06} & \textbf{-0.18} & \textbf{0.2} & \textbf{-0.16} & \textbf{0.12} & \textbf{-0.13} & \textbf{0.19} & \textbf{0.19} & \textbf{0.16} & \textbf{0.15} & \textbf{0.1} & \textbf{-0.15} & \textbf{0.05} & \textbf{-0.17} & \textbf{0.22} & \textbf{-0.02} & \textbf{-0.1} & \textbf{-0.04}\\
\multirow{3}{*}{ru}&\textsc{Opus-MT}&\textbf{15.38} & \textbf{4.07} & \textbf{52.95} & \textbf{-27.98} & \textbf{17.38} & \textbf{-24.19} & \textbf{20.45} & \textbf{11.79} & \textbf{12.51} & \textbf{12.13} & \textbf{13.6} & \textbf{-24.94} & \textbf{22.11} & \textbf{-28.52} & \textbf{41.31} & \textbf{-18.56} & \textbf{23.26} & -0.79 & -1.99 & \textbf{-28.53}\\
&deb-\textsc{Opus-MT}&\textbf{0.23} & \textbf{-0.1} & \textbf{-0.17} & \textbf{0.13} & 0.05 & 0.03 & 0.05 & -0.03 & \textbf{-0.1} & -0.04 & \textbf{-0.11} & \textbf{-0.14} & \textbf{0.09} & \textbf{0.04} & \textbf{0.13} & 0.03 & \textbf{-0.08} & \textbf{-0.13} & \textbf{-0.26} & -0.0\\
&\textsc{m2m100}&\textbf{0.08} & \textbf{-0.07} & \textbf{-0.05} & \textbf{0.06} & \textbf{-0.02} & -0.01 & \textbf{-0.02} & -0.0 & \textbf{-0.06} & \textbf{-0.07} & \textbf{-0.05} & \textbf{-0.03} & \textbf{-0.01} & 0.0 & \textbf{-0.02} & 0.0 & \textbf{-0.01} & -0.0 & 0.01 & \textbf{0.03}\\
\midrule
\multicolumn{22}{c}{\textsc{se}}\\
\midrule
\multirow{3}{*}{es}&\textsc{Opus-MT}&\textbf{12.12} & \textbf{47.0} & \textbf{83.53} & -2.45 & \textbf{20.67} & -2.44 & \textbf{27.76} & \textbf{38.98} & \textbf{49.12} & \textbf{44.16} & \textbf{37.6} & 0.23 & \textbf{31.36} & \textbf{7.95} & \textbf{42.89} & \textbf{-22.44} & \textbf{14.64} & N/A & N/A& \textbf{-65.29}\\
&deb-\textsc{Opus-MT}&\textbf{-0.08} & \textbf{-0.06} & \textbf{-0.05} & \textbf{0.05} & -0.01 & 0.0 & \textbf{-0.01} & -0.0 & \textbf{-0.04} & \textbf{-0.05} & \textbf{-0.04} & \textbf{-0.03} & \textbf{-0.01} & 0.0 & \textbf{-0.01} & \textbf{0.01} & \textbf{-0.02} & N/A & N/A& \textbf{-0.01}\\
&\textsc{m2m100}&\textbf{0.08} & \textbf{-0.07} & \textbf{-0.06} & \textbf{0.07} & -0.01 & -0.0 & \textbf{-0.02} & \textbf{-0.01} & \textbf{-0.05} & \textbf{-0.05} & \textbf{-0.04} & \textbf{-0.04} & \textbf{-0.01} & 0.0 & \textbf{-0.02} & \textbf{0.01} & \textbf{-0.02} & N/A & N/A& \textbf{-0.01}\\
\multirow{3}{*}{fr}&\textsc{Opus-MT}&\textbf{16.01} & \textbf{47.22} & \textbf{80.76} & -0.14 & \textbf{19.01} & \textbf{-8.69} & \textbf{26.55} & \textbf{31.41} & \textbf{35.76} & \textbf{33.4} & \textbf{31.98} & -1.93 & \textbf{29.87} & 2.67 & \textbf{35.97} & \textbf{-15.81} & \textbf{17.83} & N/A & N/A& \textbf{-64.01}\\
&deb-\textsc{Opus-MT}&\textbf{-0.07} & \textbf{-0.05} & \textbf{-0.04} & \textbf{0.04} & -0.0 & \textbf{0.02} & -0.01 & 0.0 & \textbf{-0.03} & \textbf{-0.04} & \textbf{-0.03} & \textbf{-0.03} & \textbf{-0.01} & \textbf{0.01} & \textbf{-0.01} & \textbf{0.01} & \textbf{-0.02} & N/A & N/A& \textbf{-0.01}\\
&\textsc{m2m100}&\textbf{0.14} & \textbf{0.38} & \textbf{0.04} & \textbf{-0.21} & \textbf{0.35} & \textbf{-0.15} & \textbf{0.23} & \textbf{-0.17} & \textbf{0.15} & \textbf{0.2} & \textbf{0.19} & \textbf{0.13} & \textbf{0.23} & \textbf{-0.16} & \textbf{0.24} & \textbf{-0.19} & \textbf{0.19} & N/A & N/A& \textbf{-0.12}\\
\multirow{3}{*}{uk}&\textsc{Opus-MT}&\textbf{-0.01} & \textbf{-0.01} & \textbf{-0.01} & \textbf{-0.03} & -0.02 & 0.0 & 0.0 & \textbf{-0.03} & \textbf{-0.03} & \textbf{-0.01} & -0.01 & -0.0 & 0.0 & \textbf{-0.01} & 0.01 & -0.0 & -0.0 & \textbf{0.02} & \textbf{0.03} & \textbf{0.01}\\
&deb-\textsc{Opus-MT}&0.0 & \textbf{0.09} & 0.02 & \textbf{-0.05} & 0.01 & -0.06 & \textbf{0.06} & -0.04 & \textbf{0.12} & \textbf{0.16} & \textbf{0.08} & \textbf{0.04} & \textbf{0.04} & \textbf{-0.02} & \textbf{0.08} & 0.02 & 0.01 & \textbf{-0.06} & \textbf{-0.1} & \textbf{-0.03}\\
&\textsc{m2m100}&\textbf{0.14} & \textbf{0.3} & \textbf{0.06} & \textbf{-0.18} & \textbf{0.2} & \textbf{-0.16} & \textbf{0.12} & \textbf{-0.13} & \textbf{0.19} & \textbf{0.19} & \textbf{0.16} & \textbf{0.15} & \textbf{0.1} & \textbf{-0.15} & \textbf{0.05} & \textbf{-0.17} & \textbf{0.22} & \textbf{-0.02} & \textbf{-0.1} & \textbf{-0.04}\\
\multirow{3}{*}{ru}&\textsc{Opus-MT}&\textbf{15.38} & \textbf{4.07} & \textbf{52.95} & \textbf{-27.98} & \textbf{17.38} & \textbf{-24.19} & \textbf{20.45} & \textbf{11.79} & \textbf{12.51} & \textbf{12.13} & \textbf{13.6} & \textbf{-24.94} & \textbf{22.11} & \textbf{-28.52} & \textbf{41.31} & \textbf{-18.56} & \textbf{23.26} & -0.79 & -1.99 & \textbf{-28.53}\\
&deb-\textsc{Opus-MT}&1.34 & -3.62 & \textbf{46.99} & \textbf{-21.71} & \textbf{-33.31} & \textbf{16.73} & \textbf{-27.33} & \textbf{19.72} & \textbf{9.4} & \textbf{7.79} & \textbf{6.6} & \textbf{9.66} & \textbf{-28.51} & \textbf{22.12} & \textbf{-24.71} & \textbf{50.74} & \textbf{-26.37} & \textbf{15.86} & -0.28 & -0.68\\
&\textsc{m2m100}&\textbf{20.01} & \textbf{18.76} & \textbf{50.38} & \textbf{-34.59} & \textbf{-11.85} & 5.91 & \textbf{-16.63} & \textbf{13.63} & \textbf{14.94} & \textbf{15.4} & \textbf{14.21} & \textbf{15.2} & \textbf{-17.46} & \textbf{14.68} & \textbf{-15.4} & \textbf{30.42} & \textbf{-10.35} & \textbf{16.48} & -0.74 & -3.23\\
\midrule
\multicolumn{22}{c}{\textsc{ge}}\\
\midrule
\multirow{3}{*}{es}&\textsc{Opus-MT}&\textbf{-0.0} & \textbf{-0.02} & \textbf{-0.01} & -0.01 & 0.01 & \textbf{-0.01} & 0.0 & \textbf{-0.02} & \textbf{-0.02} & \textbf{-0.01} & \textbf{-0.01} & \textbf{-0.01} & \textbf{0.01} & -0.0 & \textbf{0.01} & \textbf{-0.01} & -0.0 & N/A & N/A& \textbf{0.02}\\
&deb-\textsc{Opus-MT}&\textbf{-2.35} & \textbf{36.52} & \textbf{79.64} & \textbf{-58.1} & -7.06 & \textbf{18.24} & \textbf{-8.44} & \textbf{31.09} & \textbf{33.94} & \textbf{42.22} & \textbf{36.85} & \textbf{32.48} & -1.86 & \textbf{34.19} & \textbf{6.06} & \textbf{48.72} & \textbf{-29.24} & N/A & N/A& \textbf{14.06}\\
&\textsc{m2m100}&\textbf{0.19} & \textbf{0.54} & 0.02 & \textbf{-0.28} & \textbf{0.38} & \textbf{-0.27} & \textbf{0.37} & \textbf{-0.26} & \textbf{0.18} & \textbf{0.25} & \textbf{0.25} & \textbf{0.17} & \textbf{0.31} & \textbf{-0.25} & \textbf{0.31} & \textbf{-0.29} & \textbf{0.31} & N/A & N/A& \textbf{-0.17}\\
\multirow{3}{*}{fr}&\textsc{Opus-MT}&\textbf{0.06} & \textbf{-0.05} & \textbf{-0.04} & 0.01 & \textbf{0.02} & \textbf{-0.01} & 0.01 & \textbf{-0.04} & \textbf{-0.04} & \textbf{-0.03} & \textbf{-0.02} & \textbf{-0.0} & \textbf{0.01} & \textbf{-0.01} & \textbf{0.01} & \textbf{-0.02} & \textbf{-0.01} & N/A & N/A& \textbf{0.04}\\
&deb-\textsc{Opus-MT}&\textbf{-0.05} & \textbf{0.18} & \textbf{0.04} & \textbf{-0.11} & \textbf{0.08} & \textbf{-0.11} & \textbf{0.08} & \textbf{-0.11} & \textbf{0.05} & \textbf{0.12} & \textbf{0.12} & \textbf{0.06} & \textbf{0.11} & \textbf{-0.09} & \textbf{0.1} & \textbf{-0.17} & \textbf{0.11} & N/A & N/A& 0.01\\
&\textsc{m2m100}&\textbf{0.14} & \textbf{0.38} & \textbf{0.04} & \textbf{-0.21} & \textbf{0.35} & \textbf{-0.15} & \textbf{0.23} & \textbf{-0.17} & \textbf{0.15} & \textbf{0.2} & \textbf{0.19} & \textbf{0.13} & \textbf{0.23} & \textbf{-0.16} & \textbf{0.24} & \textbf{-0.19} & \textbf{0.19} & N/A & N/A& \textbf{-0.12}\\
\multirow{3}{*}{uk}&\textsc{Opus-MT}&\textbf{0.12} & \textbf{0.22} & 0.03 & \textbf{0.16} & \textbf{-0.07} & \textbf{0.09} & \textbf{-0.11} & \textbf{0.15} & \textbf{0.18} & \textbf{0.14} & \textbf{0.1} & \textbf{0.09} & \textbf{-0.1} & \textbf{0.05} & \textbf{-0.1} & \textbf{0.16} & \textbf{-0.04} & \textbf{-0.09} & \textbf{-0.05} & \textbf{-0.12}\\
&deb-\textsc{Opus-MT}&0.0 & \textbf{0.09} & 0.02 & \textbf{-0.05} & 0.01 & -0.06 & \textbf{0.06} & -0.04 & \textbf{0.12} & \textbf{0.16} & \textbf{0.08} & \textbf{0.04} & \textbf{0.04} & \textbf{-0.02} & \textbf{0.08} & 0.02 & 0.01 & \textbf{-0.06} & \textbf{-0.1} & \textbf{-0.03}\\
&\textsc{m2m100}&\textbf{0.14} & \textbf{0.3} & \textbf{0.06} & \textbf{-0.18} & \textbf{0.2} & \textbf{-0.16} & \textbf{0.12} & \textbf{-0.13} & \textbf{0.19} & \textbf{0.19} & \textbf{0.16} & \textbf{0.15} & \textbf{0.1} & \textbf{-0.15} & \textbf{0.05} & \textbf{-0.17} & \textbf{0.22} & \textbf{-0.02} & \textbf{-0.1} & \textbf{-0.04}\\
\multirow{3}{*}{ru}&\textsc{Opus-MT}&\textbf{0.09} & \textbf{0.17} & \textbf{0.05} & 0.01 & \textbf{-0.11} & \textbf{0.05} & \textbf{-0.08} & \textbf{0.08} & \textbf{0.13} & \textbf{0.11} & \textbf{0.06} & \textbf{0.04} & \textbf{-0.08} & -0.02 & \textbf{-0.06} & \textbf{0.14} & -0.01 & \textbf{-0.08} & \textbf{-0.03} & \textbf{-0.11}\\
&deb-\textsc{Opus-MT}&\textbf{-0.02} & -0.01 & -0.02 & 0.02 & \textbf{-0.05} & \textbf{-0.05} & -0.01 & -0.02 & 0.01 & 0.02 & \textbf{-0.04} & \textbf{-0.05} & \textbf{-0.03} & \textbf{-0.04} & \textbf{-0.15} & \textbf{-0.15} & \textbf{0.13} & \textbf{0.12} & \textbf{-0.05} & -0.01\\
&\textsc{m2m100}&\textbf{0.15} & \textbf{0.29} & \textbf{0.08} & \textbf{-0.18} & \textbf{0.13} & \textbf{-0.1} & \textbf{0.09} & \textbf{-0.12} & \textbf{0.16} & \textbf{0.21} & \textbf{0.15} & \textbf{0.11} & \textbf{0.07} & \textbf{-0.13} & \textbf{0.08} & \textbf{-0.05} & \textbf{0.15} & \textbf{-0.08} & \textbf{-0.1} & -0.02\\
\bottomrule
\end{tabular}
\caption{ 
\textsc{anova} results (no normalisation): single effects of bias cues (\textbf{F}eminine, \textbf{M}asculine, \textbf{S}ubject and \textbf{O}bject) on $\mathcal{H}$ (\textsc{s3e}), $\mathcal{H}$ (\textsc{se}) and $\mathcal{H}$ (\textsc{ge}). Values correspond to effect coefficients (deviations from a reference group). \textbf{Boldface} indicates statistical significance (p < 0.05). The sign of the values indicates whether the presence of the variable increases (positive) or decreases (negative) the mean $\mathcal H$ of the group containing the given variable value. Reference group is N for all columns except: `no name' for Names, `no default' for Default M, `unambiguous' for Ambiguity.
}
\label{tab:fullanova_nonorm}
\end{table*}

\clearpage

\newpage
\section{Gender Accuracy}
\label{app:genderacc}

Table~\ref{tab:genderacc} presents more fine-grained results than Table~\ref{tab:logsurprisalacc} with regard to gender accuracy, namely splitting the results by subset of the dataset. 
The results in the Ambiguous column are not meaningfully interpretable, as a single ground truth label of gender cannot capture the true desired behavior of the model, especially when the gold label for ambiguous cases is mostly `neutral', and neutral is not commonly used as grammatical gender for animate objects in the languages used in this study. 
The case of Russian, where the performance increases on the Ambiguous subset actually reflects the model choosing the masculine forms, which are tagged as `neutral' by the morphological parser due to the masculine form often being the default choice for both genders, as discussed in Section~\ref{sec:languages}.

\begin{table}[ht]
    \centering
    \scriptsize
    \resizebox{\columnwidth}{!}{%
    \begin{tabular}{lcccccc}
        \toprule
        \textbf{Lang.}& \textbf{Model} & \textbf{All} & \textbf{Pro} & \textbf{Anti} & \textbf{Unamb.} & \textbf{Amb.} \\
        \midrule
        \multirow{4}{*}{es} 
        & \textsc{Opus-MT}      & 55.20 & 67.95 & 52.10 & 67.95 & 33.96 \\
        & deb-\textsc{Opus-MT}  & 55.69 & 68.13 & 52.95 & 68.13 & 34.39 \\
        & \textsc{m2m100}    & 55.68 & 70.77 & 51.17 & 70.77 & 32.40 \\
        & \textsc{Tower+9B}    &  79.44& 86.91 & 88.08 & 86.91 & 44.02 \\
        \cmidrule{1-7}
        \multirow{4}{*}{fr} 
        & \textsc{Opus-MT}      & 52.05 & 64.27 & 46.55 & 64.27 & 37.25 \\
        & deb-\textsc{Opus-MT}  & 52.98 & 64.79 & 48.10 & 64.79 & 37.75 \\
        & \textsc{m2m100}    & 50.95 & 61.66 & 47.57 & 61.66 & 34.84 \\
        & \textsc{Tower+9B}    & 69.96 & 75.77 & 75.13 & 75.77 & 45.78 \\
        \cmidrule{1-7}
        \multirow{4}{*}{uk} 
        & \textsc{\textsc{Opus-MT}}      & 38.65 & 45.34 & 34.20 & 45.34 & 33.75 \\
        & deb-\textsc{\textsc{Opus-MT}}  & 38.95 & 46.12 & 34.15 & 46.12 & 33.74 \\
        & \textsc{m2m100}    & 40.97 & 47.76 & 36.81 & 47.76 & 35.20 \\
        & \textsc{Tower+9B}    & 55.05 & 57.58 & 58.97 & 57.58 & 40.88 \\
        \cmidrule{1-7}
        \multirow{4}{*}{ru} 
        & \textsc{\textsc{Opus-MT}}      & 39.50 & 48.57 & 33.27 & 48.57 & 33.24 \\
        & deb-\textsc{\textsc{Opus-MT}}  & 39.50 & 48.42 & 33.38 & 48.42 & 33.33 \\
        & \textsc{m2m100}    & 41.01 & 48.49 & 36.81 & 48.49 & 33.81 \\
        & \textsc{Tower+9B}    & 49.50 & 51.44 & 52.68 & 51.44 & 48.25 \\
        \bottomrule
    \end{tabular}%
    }
    \caption{Comparison of Gender Accuracy Overall, in Pro-/Anti-Stereotypical and Ambiguous Cases Across Models, on \textsc{WinoMT}.}
    \label{tab:genderacc}
\end{table}

\section{Quality Estimation with Human Translations}
\label{app:qualgender}

Table~\ref{tab:comet} presents the results of the models used in this study on the 100 human-annotated instances. For unambiguous cases we use a single reference, whereas for ambiguous cases, we calculate performance by taking the maximum \textsc{comet} score of both acceptable translations. 

\begin{table}[ht]
    \centering
    \scriptsize
    \resizebox{\columnwidth}{!}{%
    \begin{tabular}{lcccccc}
        \toprule
        \textbf{Lang.} & \textbf{Model} & \textbf{All} & \textbf{Pro} & \textbf{Anti} & \textbf{Unamb.} & \textbf{Amb.} \\
        \midrule
        \multirow{4}{*}{es}
        & \textsc{Opus-MT}      & 81.35 & 85.80 & 83.37 & 84.55 & 75.14 \\
        & deb-\textsc{Opus-MT}  & 81.31 & 85.62 & 83.43 & 84.49 & 75.14 \\
        & \textsc{m2m100}       & 79.56 & 84.44 & 81.29 & 82.82 & 73.25 \\
        & \textsc{Tower+9B}     & 81.95 & 85.79 & 84.21 & 85.00 & 76.08 \\
        \cmidrule{1-7}
        \multirow{4}{*}{fr}
        & \textsc{Opus-MT}      & 77.63 & 82.23 & 81.16 & 81.66 & 70.47 \\
        & deb-\textsc{Opus-MT}  & 77.69 & 82.41 & 80.88 & 81.60 & 70.73 \\
        & \textsc{m2m100}       & 76.24 & 80.69 & 78.65 & 79.61 & 70.25 \\
        & \textsc{Tower+9B}     & 81.51 & 85.63 & 83.76 & 84.70 & 75.95 \\
        \cmidrule{1-7}
        \multirow{4}{*}{uk}
        & \textsc{Opus-MT}      & 80.56 & 85.57 & 82.73 & 84.10 & 73.69 \\
        & deb-\textsc{Opus-MT}  & 80.19 & 84.85 & 82.13 & 83.45 & 73.86 \\
        & \textsc{m2m100}       & 81.27 & 85.89 & 84.09 & 84.96 & 74.10 \\
        & \textsc{Tower+9B}     & 85.51 & 90.18 & 88.05 & 89.12 & 78.57 \\
        \cmidrule{1-7}
        \multirow{4}{*}{ru}
        & \textsc{Opus-MT}      & 82.53 & 86.20 & 84.99 & 85.58 & 76.86 \\
        & deb-\textsc{Opus-MT}  & 82.76 & 86.46 & 85.27 & 85.85 & 77.02 \\
        & \textsc{m2m100}       & 81.71 & 86.39 & 84.06 & 85.21 & 75.21 \\
        & \textsc{Tower+9B}     & 85.06 & 89.46 & 87.27 & 88.37 & 78.96 \\
        \bottomrule
    \end{tabular}%
    }
    \caption{Comparison of \textsc{comet} Scores Overall, in Pro-/Anti-Stereotypical and Ambiguous Cases Across Models, on the 100 manually translated sentences.}
    \label{tab:comet}
\end{table}

\clearpage
\section{Log Probability and Surprisal Scores}
\label{app:logsurprisal}

Table~\ref{tab:logsurprisal} presents the Log probability and surprisal scores, as well as their relative differences between the Correct and Incorrect translations of the Unambiguous instances in \textsc{WinoMT}. 

\begin{table*}[ht]
\centering
\begin{adjustbox}{max width=\textwidth}
\begin{tabular}{ll|cc|cc|cc|cc}
\toprule
Language & Model & LogProb (Correct) & LogProb (Incorrect) & \textsc{s3e} $I$ (Correct) & \textsc{s3e} $I$ (Incorrect) & \textsc{se} $I$ (Correct) & \textsc{se} $I$ (Incorrect) & \textsc{ge} $I$ (Correct) & \textsc{ge} $I$ (Incorrect) \\
\midrule
\multirow{3}{*}{ES} 
    & \textsc{Opus-MT} & -149.7  & -149.78 & 7.83   & 8.88   & 0.3  & 0.35 & 0.33 & 0.3 \\
    & deb-\textsc{Opus-MT} & -149.19 & -149.01 & 8.08   & 9.16   & 0.29 & 0.31 & 0.35 & 0.33 \\
    & \textsc{m2m100} & -226.61 & -227.29 & 23.61  & 26.03  & 0.41 & 0.4  & 0.42 & 0.43 \\
    & \textsc{Tower+9B} & -586.29 & -586.72 & 15.25 & 19.80 & 0.97 & 1.19 & 0.51 & 0.51 \\
\midrule
\multirow{3}{*}{FR} 
    & \textsc{Opus-MT} & -197.1  & -195.11 & 9.18   & 9.89   & 0.73 & 0.72 & 0.24 & 0.29 \\
    & deb-\textsc{Opus-MT} & -196.98 & -195.09 & 9.18   & 9.85   & 0.48 & 0.52 & 0.26 & 0.33 \\
    & \textsc{m2m100} & -283.91 & -281.71 & 186.42 & 194.52 & 0.49 & 0.4  & 0.43 & 0.48 \\
        & \textsc{Tower+9B} & -661.37 & -677.52 & 15.62 & 19.79 & 1.24 & 1.33 & 0.78 &  0.78\\
\midrule
\multirow{3}{*}{UK} 
    & \textsc{Opus-MT} & -161.98 & -161.14 & 147.6  & 152.15 & 0.6   & 0.54 & 0.22 & 0.22 \\
    & \textsc{Opus-MT}-debiased & -161.46 & -160.68 & 150.52 & 153.76 & 0.49  & 0.47 & 0.23 & 0.23 \\
    & \textsc{m2m100} & -241.0  & -241.49 & 204.72 & 211.9 & 0.28  & 0.24 & 0.23 & 0.25 \\
    & \textsc{Tower+9B} & -455.91 & -462.70 & 16.84 & 18.81 & 1.5 & 1.59 & 1.62 & 1.47 \\
\midrule
\multirow{3}{*}{RU} 
    & \textsc{Opus-MT} & -170.72 & -170.9  & 32.14  & 33.15  & 0.38 & 0.43 & 0.08 & 0.19 \\
    & deb-\textsc{Opus-MT} & -170.58 & -170.75 & 32.31  & 33.27  & 0.32 & 0.37 & 0.06 & 0.17 \\
    & \textsc{m2m100} & -220.25 & -220.78 & 218.11 & 219.06 & 0.45 & 0.4  & 0.16 & 0.3 \\
    & \textsc{Tower+9B} & -499.57 & -508.94 & 27.34 & 29.75 & 1.95 & 2.11 & 1.0 & 0.78 \\
\bottomrule
\end{tabular}
\end{adjustbox}
\caption{Log Probability and Surprisal Measures across Models and Languages}
\label{tab:logsurprisal}
\end{table*}

\section{Results on the \textsc{mGeNTe} dataset}
\label{app:mgenteresults}

Table~\ref{tab:logsurprisalacc2} presents the results of the experiments on the \textsc{mGeNTe} dataset. 

\begin{table}[ht]
\centering
\begin{adjustbox}{max width=\columnwidth}
\begin{tabular}{ll|c|c|c|c}
\toprule
\textbf{Lang.} & \textbf{Model} & \textbf{Gender Acc} & \textbf{$\Delta \text{Log prob}$} & \textbf{$\Delta I$ (\textsc{s3e})} & \textbf{\textsc{comet}} \\
\midrule
\multirow{3}{*}{ES} 
& \textsc{Opus-MT}         & 60.40 &  -0.06 & -1.92  & 84.90  \\
& deb-\textsc{Opus-MT}     & 60.67 &  -0.06 & -1.91  &  84.86 \\
& \textsc{m2m100}          & 60.53 &  -0.05 & 1.73  &  84.53 \\
& \textsc{Tower+9B}        & 67.47 & -0.04  & 4.53  &  86.88 \\
\midrule
\multirow{3}{*}{DE} 
& \textsc{Opus-MT}         & 69.20 &  -0.06 & -1.45  &  81.30 \\
& deb-\textsc{Opus-MT}     & 68.93 &  -0.06 & -1.43 & 81.24 \\
& \textsc{m2m100}          & 68.53 &  -0.06 &  1.96 & 82.29\\
& \textsc{Tower+9B}        & 79.87 &  -0.04 & 2.21  &  85.33 \\
\midrule
\multirow{3}{*}{IT} 
& \textsc{Opus-MT}         & 64.40 &  -0.04 &  -1.45 & 84.65  \\
& deb-\textsc{Opus-MT}     & 64.40 &  -0.04 &  -1.46 & 84.58  \\
& \textsc{m2m100}          & 66.00 &  -0.04 &  1.96 &  85.04 \\
& \textsc{Tower+9B}        & 74.00 & -0.03  & 2.21  &  87.78 \\
\midrule
\multirow{3}{*}{EL} 
& \textsc{Opus-MT}         & 65.73 &  -0.05 &  -1.43 &  78.41 \\
& deb-\textsc{Opus-MT}     & 65.87 &  -0.05 &  -1.41 & 77.90  \\
& \textsc{m2m100}          & 64.27 &  -0.08 & 1.96  &  78.58 \\
& \textsc{Tower+9B}        & 64.11 &  -0.02 &  2.18 &  66.23 \\ 
\bottomrule
\end{tabular}
\end{adjustbox}
\caption{Gender Accuracy, $\Delta$ Log Probability, and $\Delta I$ (\textsc{s3e}) on Unambiguous instances of \textsc{mGeNTe}, \textsc{comet} scores on WMT test sets.}
\label{tab:logsurprisalacc2}
\end{table}

There is no correlation between $\Delta I$ scores and gender accuracy on the \textsc{mGeNTe} dataset (see Table~\ref{tab:kendalpearson2}).

\begin{table}[ht]
\centering
\begin{tabular}{llcc}
\toprule
\textbf{Correlation} & \textbf{Metric} & \textbf{Statistic} & \textbf{p-value} \\
\midrule
\multirow{4}{*}{Spearman} 
  &  $\Delta I$ (\textsc{s3e})& -0.40 & 0.12 \\
  &  $\Delta I$ (\textsc{se}) & -0.35 & 0.19 \\
  &  $\Delta I$ (\textsc{ge}) & 0.44 & 0.09 \\ 
  & $\Delta \text{Log prob}$ & -0.11 & 0.70 \\
\midrule
\multirow{4}{*}{Kendall} 
  &  $\Delta I$ (\textsc{s3e}) & -0.28 & 0.14 \\
  &  $\Delta I$ (\textsc{se}) & -0.18 & 0.34 \\
  &  $\Delta I$ (\textsc{ge}) & 0.30 & 0.11 \\
  & $\Delta \text{Log prob}$ & -0.10 & 0.59 \\
\bottomrule
\end{tabular}
\caption{Spearman and Kendall correlations between $\Delta I$ under different uncertainty metrics and Log Probabilities on the one hand, and gender accuracy on the other on the \textsc{mGeNTe} dataset. No statistically significant correlations present.}
\label{tab:kendalpearson2}
\end{table}

Table~\ref{tab:s3e_entropy_structured_sorted2} presents the entropy (\textsc{s3e}) scores on the \textsc{mGeNTe} dataset. The best performing model (\textsc{Tower+9B}) shows the lowest $\Delta\mathcal{H}$ across languages, indicating the least bias. 

\begin{table}[ht]
\centering
\footnotesize
\begin{tabular}{@{}l l | ccc@{}}
\toprule
Lang. & Model & Unamb & Amb & $\Delta\mathcal{H}$\\ 
\midrule
\multirow{3}{*}{ES} 
    & \textsc{Opus-MT} & 2.53 & 2.57 & \textbf{-0.02} \\
    & deb-\textsc{Opus-MT}  & 2.45 & 2.47 & \textbf{-0.01}\\
    & \textsc{m2m100} &  2.41 & 2.28 & 0.06 \\
    & \textsc{Tower+9B} & 3.77 & 4.08 & \textbf{-0.08} \\
\midrule
\multirow{3}{*}{DE} 
    & \textsc{Opus-MT} & 2.90 & 2.87 & 0.01 \\
    & deb-\textsc{Opus-MT}  & 2.92 & 2.90 & 0.01  \\
    & \textsc{m2m100} & 2.72  & 2.90 & -0.06 \\
    & \textsc{Tower+9B} & 3.36 & 4.35 & \textbf{-0.25} \\
\midrule
\multirow{3}{*}{IT} 
    & \textsc{Opus-MT} & 2.89 & 2.78 & 0.04 \\
    & deb-\textsc{Opus-MT}  & 2.17 & 1.79 & 0.19 \\
    & \textsc{m2m100} &  2.91 & 2.63 & 0.10 \\
    & \textsc{Tower+9B} & 3.77 & 4.08 & \textbf{-0.08} \\
\midrule
\multirow{3}{*}{EL} 
    & \textsc{Opus-MT} & 2.80 & 2.67 & 0.05 \\
    & deb-\textsc{Opus-MT}  & 2.75 & 2.60 & 0.05 \\
    & \textsc{m2m100} & 3.19 & 2.88  & 0.10 \\
    & \textsc{Tower+9B} & 6.74 & 6.99 & -0.04 \\
\bottomrule
\end{tabular}
\caption{Unambiguous and Ambiguous $\mathcal{H}$ (\textsc{s3e}) on \textsc{mGeNTe}. Statistically significant correlations between the $\Delta\mathcal{H}$ and \textsc{comet} scores at the instance level are presented in \textbf{bold}.}
\label{tab:s3e_entropy_structured_sorted2}
\end{table}

\section{Rankings by Different Metrics}
\label{app:rankings}

Table~\ref{tab:model_rankings} presents the rankings of models according to the metrics employed in this study. 

\begin{table}[ht]
\centering
\footnotesize
\renewcommand{\arraystretch}{1.1}
\begin{tabular}{l l}
\toprule
\textbf{Gender Acc} & \textbf{Delta S} \\
\midrule
\textsc{tower}-ES                 & \textsc{tower}-ES \\
\textsc{m2m100}-ES                & \textsc{tower}-FR \\
\textsc{tower}-FR                 & deb-\textsc{Opus-MT}-ES \\
deb-\textsc{Opus-MT}-ES           & \textsc{m2m100}-ES \\
\textsc{Opus-MT}-ES               & \textsc{tower}-UK \\
deb-\textsc{Opus-MT}-FR           & \textsc{Opus-MT}-ES \\
\textsc{Opus-MT}-FR               & \textsc{tower}-RU \\
\textsc{m2m100}-FR                & deb-\textsc{Opus-MT}-FR \\
\textsc{tower}-UK                 & \textsc{m2m100}-FR \\
\textsc{tower}-RU                 & \textsc{Opus-MT}-FR \\
\textsc{Opus-MT}-UK               & \textsc{Opus-MT}-RU \\
\textsc{m2m100}-UK                & deb-\textsc{Opus-MT}-UK \\
deb-\textsc{Opus-MT}-UK           & deb-\textsc{Opus-MT}-RU \\
\textsc{m2m100}-RU                & \textsc{m2m100}-UK \\
deb-\textsc{Opus-MT}-RU           & \textsc{m2m100}-RU \\
\textsc{Opus-MT}-RU               & \textsc{Opus-MT}-UK \\
\bottomrule
\end{tabular}
\caption{Model rankings for gender accuracy (higher is better) and \textsc{s3e} delta (lower is better) on the \textsc{WinoMT} dataset.}
\label{tab:model_rankings}
\end{table}

\newpage
\section{Rank Correlation}
\label{app:kendalpearson}

Table~\ref{tab:kendalpearson} presents the correlation scores between $\Delta I$ and $\Delta$ Log Probabilies on the one hand, and gender accuracy scores on the other. 

\begin{table}[ht]
\centering
\begin{tabular}{llcc}
\toprule
\textbf{Correlation} & \textbf{Metric} & \textbf{Statistic} & \textbf{p-value} \\
\midrule
\multirow{4}{*}{Spearman} 
  &  $\Delta I$ (\textsc{s3e})& \textbf{-0.84} & \textbf{0.00} \\
  &  $\Delta I$ (\textsc{se}) & -0.42 & 0.10 \\
  &  $\Delta I$ (\textsc{ge}) & 0.29 & 0.27 \\ 
  & $\Delta \text{Log prob}$ & 0.00 & 1.00 \\
\midrule
\multirow{4}{*}{Kendall} 
  &  $\Delta I$ (\textsc{s3e}) & \textbf{-0.63} & \textbf{0.00} \\
  &  $\Delta I$ (\textsc{se}) & -0.30 & 0.12 \\
  &  $\Delta I$ (\textsc{ge}) & 0.21 & 0.27 \\
  & $\Delta \text{Log prob}$ & -0.03 & 0.89 \\
\bottomrule
\end{tabular}
\caption{Spearman and Kendall correlations between $\Delta I$ under different uncertainty metrics and Log Probabilities on the one hand, and gender accuracy on the other, for the \textsc{WinoMT} dataset. Statistically significant correlations (p $<$ 0.05) are in \textbf{bold}.}
\label{tab:kendalpearson}
\end{table}

\newpage
\section{Entropy Scores}
\label{app:entropy}

Table~\ref{tab:entropy_deltas} presents the $\mathcal H$ scores and their relative differences between the unambiguous and ambiguous settings for different UQ metrics used in this study. 

\begin{table*}[ht]
\centering
\begin{adjustbox}{max width=\textwidth}
\begin{tabular}{ll|ccc|ccc|ccc}
\toprule
\multirow{2}{*}{Language} & \multirow{2}{*}{Model} & \multicolumn{3}{c|}{\textsc{s3e}} & \multicolumn{3}{c|}{\textsc{se}} & \multicolumn{3}{c}{\textsc{ge}} \\
&& Unamb. & Amb. & $\Delta\mathcal{H}$
& Unamb. & Amb. & $\Delta\mathcal{H}$ 
& Unamb. & Amb. & $\Delta\mathcal{H}$ \\
\midrule
\multirow{3}{*}{ES} 
    & \textsc{Opus-MT} & 1.23 & 1.12 & 0.09 & 0.33 & 0.22 & 0.33 & 0.21 & 0.13 & 0.38 \\
    & deb-\textsc{Opus-MT} & 0.97 & 0.89 & 0.08 & 0.41 & 0.25 & 0.39 & 0.24 & 0.16 & 0.33 \\
    & \textsc{m2m100} & 1.79 & 1.45 & 0.19 & 0.46 & 0.17 & 0.63 & 0.25 & 0.09 & 0.64 \\
\midrule
\multirow{3}{*}{FR} 
    & \textsc{Opus-MT} & 1.79 & 1.43 & 0.20 & 0.57 & 0.40 & 0.30 & 0.23 & 0.08 & 0.65 \\
    & deb-\textsc{Opus-MT} & 1.21 & 1.08 & 0.11 & 0.64 & 0.50 & 0.22 & 0.23 & 0.09 & 0.61 \\
    & \textsc{m2m100} & 3.22 & 2.78 & 0.14 & 0.56 & 0.29 & 0.48 & 0.25 & 0.17 & 0.32 \\
\midrule
\multirow{3}{*}{UK} 
    & \textsc{Opus-MT} & 1.96 & 2.16 & -0.10 & 0.40 & 0.39 & 0.03 & 0.20 & 0.14 & 0.30 \\
    & deb-\textsc{Opus-MT} & 1.98 & 2.15 & -0.09 & 0.44 & 0.41 & 0.07 & 0.22 & 0.15 & 0.32 \\
    & \textsc{m2m100} & 2.05 & 2.28 & -0.11 & 0.37 & 0.41 & -0.11 & 0.18 & 0.16 & 0.11 \\
\midrule
\multirow{3}{*}{RU} 
    & \textsc{Opus-MT} & 1.56 & 1.68 & -0.08 & 0.38 & 0.32 & 0.16 & 0.12 & 0.03 & 0.75 \\
    & deb-\textsc{Opus-MT} & 1.05 & 0.97 & 0.08 & 0.43 & 0.42 & 0.02 & 0.12 & 0.06 & 0.50 \\
    & \textsc{m2m100} & 1.83 & 2.29 & -0.25 & 0.50 & 0.29 & 0.42 & 0.15 & 0.10 & 0.33 \\
\bottomrule
\end{tabular}
\end{adjustbox}
\caption{$\mathcal H$ scores across models and languages, with relative differences ($\Delta \mathcal{H}$) between unambiguous and ambiguous conditions on the \textsc{WinoMT} dataset.}
\label{tab:entropy_deltas}
\end{table*}

\end{document}